\documentclass{article}

\PassOptionsToPackage{authoryear, sort}{natbib}

 \usepackage[preprint]{neurips_2024}

\usepackage{import}%
\usepackage{graphicx}%
\usepackage{comment}%
\usepackage{multirow}%
\usepackage{amsmath,amssymb,amsfonts}%
\usepackage{amsthm}%
\usepackage{mathrsfs}%
\usepackage[title]{appendix}%
\usepackage[usenames, dvipsnames]{xcolor}%
\usepackage{textcomp}%
\usepackage{manyfoot}%
\usepackage{booktabs}%
\usepackage{array}
\usepackage{algorithm}%
\usepackage{algorithmicx}%
\usepackage{algpseudocode}%
\usepackage{listings}%
\usepackage{longtable}%
\usepackage[most]{tcolorbox}%
\usepackage{booktabs}
\usepackage{enumerate}
\usepackage{url}

\usepackage{hyperref}
\usepackage{tikz}
\usepackage{pgfplots}
\usepackage{array} % required for text wrapping in tables
\usetikzlibrary{shapes, arrows, positioning, fit}
\usepackage{tabularx}
\usepackage{placeins}

\hypersetup{
    colorlinks=true,
    allcolors=[rgb]{0 0 0.5}
    }

\title{From Local to Global: A GraphRAG Approach to Query-Focused Summarization}

\author{%
    Darren Edge\textsuperscript{1†} 
    \And Ha Trinh\textsuperscript{1†} 
    \And Newman Cheng\textsuperscript{2} 
    \And Joshua Bradley\textsuperscript{2} 
    \And Alex Chao\textsuperscript{3} 
    \And Apurva Mody\textsuperscript{3} 
    \And Steven Truitt\textsuperscript{2} 
    \And Dasha Metropolitansky\textsuperscript{1}\\ 
    \And Robert Osazuwa Ness\textsuperscript{1}
    \And Jonathan Larson\textsuperscript{1} \\\\
  \textsuperscript{1}Microsoft Research\\
  \textsuperscript{2}Microsoft Strategic Missions and Technologies\\
  \textsuperscript{3}Microsoft Office of the CTO\\
  \\ 
  \texttt{\{daedge,trinhha,newmancheng,joshbradley,achao,moapurva,} \\
  \small
\texttt{steventruitt,dasham,robertness,jolarso\}@microsoft.com} \\ \\
\small
\textsuperscript{†}These authors contributed equally to this work 
}
\pgfplotsset{compat=1.18}
\begin{document}

\maketitle

\begin{abstract}
The use of retrieval-augmented generation (RAG) to retrieve relevant information from an external knowledge source enables large language models (LLMs) to answer questions over private and/or previously unseen document collections.
However, RAG fails on global questions directed at an entire text corpus, such as ``What are the main themes in the dataset?'', since this is inherently a query-focused summarization (QFS) task, rather than an explicit retrieval task.
Prior QFS methods, meanwhile, do not scale to the quantities of text indexed by typical RAG systems. 
To combine the strengths of these contrasting methods, we propose \textit{GraphRAG}, a graph-based approach to question answering over private text corpora that scales with both the generality of user questions and the quantity of source text.
Our approach uses an LLM to build a graph index in two stages: first, to derive an entity knowledge graph from the source documents, then to pregenerate community summaries for all groups of closely related entities.
Given a question, each community summary is used to generate a partial response, before all partial responses are again summarized in a final response to the user.
For a class of global sensemaking questions over datasets in the 1 million token range, we show that GraphRAG leads to substantial improvements over a conventional RAG baseline for both the comprehensiveness and diversity of generated answers.

\end{abstract}

\section{Introduction}

Retrieval augmented generation (RAG)~\citep{lewis2020retrieval} is an established approach to using LLMs to answer queries based on data that is too large to contain in a language model's \emph{context window}, meaning the maximum number of \emph{tokens} (units of text) that can be processed by the LLM at once ~\citep{liu-etal:2023:tacl, kuratov2024search}.
In the canonical RAG setup, the system has access to a large external corpus of text records and retrieves a subset of records that are individually relevant to the query and collectively small enough to fit into the context window of the LLM. The LLM then generates a response based on both the query and the retrieved records \citep{dang2006duc, yao2017recent, baumel2018query, laskar2020query}.
This conventional approach, which we collectively call \emph{vector RAG}, works well for queries that can be answered with information localized within a small set of records.
However, vector RAG approaches do not support \textit{sensemaking} queries, meaning queries that require global understanding of the entire dataset, such as "\emph{What are the key trends in how scientific discoveries are influenced by interdisciplinary research over the past decade?}"

\textit{Sensemaking} tasks require reasoning over ``\emph{connections
(which can be among people, places, and events) in order to anticipate their trajectories and act effectively}''~\citep{klein2006amaking}.
LLMs such as GPT~\citep{brown2020language, achiam2023gpt}, Llama~\citep{touvron2023llama}, and Gemini~\citep{team2023gemini} excel at sensemaking in complex domains like scientific discovery~\citep{ai4science2023impact} and intelligence analysis~\citep{ranade2023fabula}.
Given a sensemaking query and a text with an implicit and interconnected set of concepts, an LLM can generate a summary that answers the query.
The challenge, however, arises when the volume of data requires a RAG approach, since vector RAG approaches are unable to support sensemaking over an entire corpus.

In this paper, we present \textbf{GraphRAG} -- a graph-based RAG approach that enables sensemaking over the entirety of a large text corpus.
GraphRAG first uses an LLM to construct a knowledge graph, where nodes correspond to key entities in the corpus and edges represent relationships between those entities. 
Next, it partitions the graph into a hierarchy of communities of closely related entities, before using an LLM to generate community-level summaries. These summaries are generated in a bottom-up manner following the hierarchical structure of extracted communities, with summaries at higher levels of the hierarchy recursively incorporating lower-level summaries.
Together, these community summaries provide global descriptions and insights over the corpus.
Finally, GraphRAG answers queries through map-reduce processing of community summaries; in the map step, the summaries are used to provide partial answers to the query independently and in parallel, then in the reduce step, the partial answers are combined and used to generate a final global answer.

The GraphRAG method and its ability to perform global sensemaking over an entire corpus form the main contribution of this work. To demonstrate this ability, we developed a novel application of the LLM-as-a-judge technique ~\citep{zheng2024judging} suitable for questions targeting broad issues and themes where there is no ground-truth answer.
This approach first uses one LLM to generate a diverse set of global sensemaking questions based on corpus-specific use cases, before using a second LLM to judge the answers of two different RAG systems using predefined criteria (defined in \autoref{sec:criteria}).
We use this approach to compare GraphRAG to vector RAG on two representative real-world text datasets.
Results show GraphRAG strongly outperforms vector RAG when using GPT-4 as the LLM.

GraphRAG is available as open-source software at \url{https://github.com/microsoft/graphrag}.
In addition, versions of the GraphRAG approach are also available as extensions to multiple open-source libraries, including LangChain~\citep{LangchainGraphRAG}, LlamaIndex~\citep{LlamaIndexGraphRAG}, NebulaGraph~\citep{NebulaGraph}, and Neo4J~\citep{Neo4jGraphRAG}.

\section{Background}

\subsection{RAG Approaches and Systems}

RAG generally refers to any system where a user query is used to retrieve relevant information from external data sources, whereupon this information is incorporated into the generation of a response to the query by an LLM (or other generative AI model, such as a multi-media model).
The query and retrieved records populate a prompt template, which is then passed to the LLM~\citep{ram2023context}.
RAG is ideal when the total number of records in a data source is too large to include in a single prompt to the LLM, i.e. the amount of text in the data source exceeds the LLM's context window.   

In canonical RAG approaches, the retrieval process returns a set number of records that are semantically similar to the query and the generated answer uses only the information in those retrieved records.
A common approach to conventional RAG is to use text embeddings, retrieving records closest to the query in vector space where closeness corresponds to semantic similarity \citep{gao2023retrieval}.
While some RAG approaches may use alternative retrieval mechanisms, we collectively refer to the family of conventional approaches as \emph{vector RAG}.
GraphRAG contrasts with vector RAG in its ability to answer queries that require global sensemaking over the entire data corpus.

GraphRAG builds upon prior work on advanced RAG strategies.
GraphRAG leverages summaries over large sections of the data source as a form of "self-memory" (described in~\citealt{cheng2024lift}), which are later used to answer queries as in~\citealt{mao2020generation}). These summaries are generated in parallel and iteratively aggregated into global summaries, similar to prior techniques~\citep{shao2023enhancing, wang2024feb4rag, su2020caire, feng2023retrieval, trivedi2022interleaving, khattab2022demonstrate, gao2023retrieval}.
In particular, GraphRAG is similar to other approaches that use hierarchical indexing to create summaries (similar to~\citealt{kim2023tree, sarthi2024raptor}).
GraphRAG contrasts with these approaches by generating a graph index from the source data, then applying graph-based community detection to create a thematic partitioning of the data. 

\subsection{Using Knowledge Graphs with LLMs and RAG}
\label{sec:graphlit}

Approaches to knowledge graph extraction from natural language text corpora include rule-matching, statistical pattern recognition, clustering, and embeddings~\citep{yates-etal-2007-textrunner, mooney, kimprob, knowitall}.
GraphRAG falls into a more recent body of research that use of LLMs for knowledge graph extraction~\citep{melnyk2022knowledge, representationlearning, openai2023chatgpt, ban2023query, zhang2024causal, Trajanoska2023EnhancingKG, yao2023exploring, yates-etal-2007-textrunner}.
It also adds to a growing body of RAG approaches that use a knowledge graph as an index~\citep{gao2023retrieval}.
Some techniques use subgraphs, elements of the graph, or properties of the graph structure directly in the prompt ~\citep{baek2023knowledge, he2024g, zhang2023graph} or as factual grounding for generated outputs~\citep{kang2023knowledge, ranade2023fabula}.
Other techniques~\citep{wang2023knowledge} use the knowledge graph to enhance retrieval, where at query time an LLM-based agent dynamically traverses a graph with nodes representing document elements (e.g., passages, tables) and edges encoding lexical and semantical similarity or structural relationships.
GraphRAG contrasts with these approaches by focusing on a previously unexplored quality of graphs in this context: their inherent \emph{modularity}~\citep{newman2006modularity} and the ability to partition graphs into nested modular communities of closely related nodes (e.g., Louvain,~\citealt{blondel2008fast}; Leiden,~\citealt{Traag2019Leiden}).
Specifically, GraphRAG recursively creates increasingly global summaries by using the LLM to create summaries spanning this community hierarchy.

\subsection{Adaptive benchmarking for RAG Evaluation}
Many benchmark datasets for open-domain question answering exist, including HotPotQA~\citep{yang2018hotpotqa}, MultiHop-RAG~\citep{tang2024multihop}, and MT-Bench~\citep{zheng2024judging}.
However, these benchmarks are oriented towards vector RAG performance, i.e., they evaluate performance on explicit fact retrieval. 
In this work, we propose an approach for generating a set of questions for evaluating global sensemaking over the entirety of the corpus.
Our approach is related to LLM methods that use a corpus to generate questions whose answers would be summaries of the corpus, such as in~\cite{xu2021text}.
However, in order to produce a fair evaluation, our method avoids generating the questions directly from the corpus itself (as an alternative implementation, one can use a subset of the corpus held out from subsequent graph extraction and answer evaluation steps).

\emph{Adaptive benchmarking} refers to the process of dynamically generating evaluation benchmarks tailored to specific domains or use cases.
Recent work has used LLMs for adaptive benchmarking to ensure relevance, diversity, and alignment with the target application or task \citep{yuan2024s, zhang2024darg}.
In this work, we propose an adaptive benchmarking approach to generating global sensemaking queries for the LLM.
Our approach builds on prior work in LLM-based persona generation, where the LLM is used to generate diverse and authentic sets of personas \citep{salminen2024deus, shin2024understanding, kosinski2024evaluating}.
Our adaptive benchmarking procedure uses persona generation to create queries that are representative of real-world RAG system usage.
Specifically, our approach uses the LLM to infer the potential users would use the RAG system and their use cases, which guide the generation of corpus-specific sensemaking queries.

\subsection{RAG evaluation criteria}
Our evaluation relies on the LLM to evaluate how well the RAG system answers the generated questions.
Prior work has shown LLMs to be good evaluators of natural language generation, including work where LLMs evaluations were competitive with human evaluations~\citep{wang2023chatgpt,zheng2024judging}.
Some prior work proposes criteria for having LLMs quantify the quality of generated texts such as ``fluency" ~\citep{wang2023chatgpt}
Some of these criteria are generic to vector RAG systems and not relevant to global sensemaking, such as ``context relevance", ``faithfulness", and ``answer relevance" (RAGAS,~\citealt{es2023ragas}).
Lacking a gold standard for evaluation, one can quantify relative performance for a given criterion by prompting the LLM to compare generations from two different competing models (LLM-as-a-judge,~\citep{zheng2024judging}).
In this work, we design criteria for evaluating RAG-generated answers to global sensemaking questions and evaluate our results using the comparative approach.
We also validate results using statistics derived from LLM-extracted statements of verifiable facts, or ``claims."

\section{Methods}
\subsection{GraphRAG Workflow}

\autoref{fig:pipeline} illustrates the high-level data flow of the GraphRAG approach  and pipeline.
In this section, we describe the key design parameters, techniques, and implementation details for each step.

\begin{figure}
    \centering
    \begin{tikzpicture}[->,>=stealth',shorten >=1pt,right,node distance=1.6cm,semithick,
        index/.style={
        rectangle,
        draw=black,
        thick,
        fill=blue!10,
        text width=4cm,
        align=center,
        rounded corners,
        minimum height=2em
        },
        query/.style={
        rectangle,
        draw=black,
        thick,
        fill=ForestGreen!15,
        text width=4cm,
        align=center,
        rounded corners,
        minimum height=2em
    }]
      \node[index] (Source Text Documents) {Source Documents};

      \node[index] (Source Text Chunks) [below of=Source Text Documents] {Text Chunks};
      \path (Source Text Documents) edge node[text width=2cm] {\small text extraction and chunking} (Source Text Chunks);
      
      \node[index] (Graph Element Instances) [below of=Source Text Chunks] {Entities \& Relationships};
      \path (Source Text Chunks) edge node[text width=2.5cm] {\small domain-tailored summarization} (Graph Element Instances);
      
      \node[index] (Graph Element Summaries) [below of=Graph Element Instances] {Knowledge Graph};
      \path (Graph Element Instances) edge node[text width=2.5cm] {\small domain-tailored summarization} (Graph Element Summaries);
      
      \node[index] (Graph Community Structures) [right=3cm of Graph Element Summaries] {Graph Communities};
      \path (Graph Element Summaries) edge node[text width=2cm,auto,align=center] {\small community detection} (Graph Community Structures);
      
      \node[index]  (Graph Community Summaries) [above of=Graph Community Structures] {Community Summaries};
      \path (Graph Community Structures) edge node[text width=2.5cm] {\small domain-tailored summarization} (Graph Community Summaries);
      
      \node[query]  (Community Query Summaries) [above of=Graph Community Summaries] {Community Answers};
      \path (Graph Community Summaries) edge node[text width=2.5cm] {\small query-focused summarization} (Community Query Summaries);
      
      \node[query] (Global Query Summary) [above of=Community Query Summaries] {Global Answer};
      \path (Community Query Summaries) edge node[text width=2.5cm] {\small query-focused summarization} (Global Query Summary);

      \node[index, font=\itshape] (Indexing Time) [below of=Graph Element Summaries] {Indexing Time};

      \node[query, font=\itshape] (Query Time) [below of=Graph Community Structures] {Query Time};

      \node (Stage) [right=0.3cm of Indexing Time, font=\bfseries] {Pipeline Stage};
      
    \end{tikzpicture}
    \caption{
    Graph RAG pipeline using an LLM-derived graph index of source document text.
    This graph index spans nodes (e.g., entities), edges (e.g., relationships), and covariates (e.g., claims) that have been detected, extracted, and summarized by LLM prompts tailored to the domain of the dataset.
    Community detection (e.g., Leiden,~\citeauthor{Traag2019Leiden}, \citeyear{Traag2019Leiden}) is used to partition the graph index into groups of elements (nodes, edges, covariates) that the LLM can summarize in parallel at both indexing time and query time.
    The ``global answer'' to a given query is produced using a final round of query-focused summarization over all community summaries reporting relevance to that query.
    }
    \label{fig:pipeline}
\end{figure}

\subsubsection{Source Documents \texorpdfstring{\texorpdfstring{$\rightarrow$}{-->}}{-->} Text Chunks}

To start, the documents in the corpus are split into text chunks.
The LLM extracts information from each chunk for downstream processing.
Selecting the size of the chunk is a fundamental design decision; longer text chunks require fewer LLM calls for such extraction (which reduces cost) but suffer from degraded recall of information that appears early in the chunk~\citep{liu-etal:2023:tacl, kuratov2024search}.
See \autoref{app:entity} for prompts and examples of the recall-precision trade-offs.

\subsubsection{Text Chunks \texorpdfstring{$\rightarrow$}{-->} Entities \& Relationships}

In this step, the LLM is prompted to extract instances of important \emph{entities} and the \emph{relationships} between the entities from a given chunk. 
Additionally, the LLM generates short descriptions for the entities and relationships. To illustrate, suppose a chunk contained the following text:

\begin{quote}
NeoChip's (NC) shares surged in their first week of trading on the NewTech Exchange.
However, market analysts caution that the chipmaker's public debut may not reflect trends for other technology IPOs.
NeoChip, previously a private entity, was acquired by Quantum Systems in 2016.
The innovative semiconductor firm specializes in low-power processors for wearables and IoT devices.
\end{quote}

The LLM is prompted such that it extracts the following:
\begin{itemize}
    \item The entity \texttt{NeoChip}, with description ``NeoChip is a publicly traded company specializing in low-power processors for wearables and IoT devices."
    \item The entity \texttt{Quantum Systems}, with description ``Quantum Systems is a firm that previously owned NeoChip."
    \item A relationship between \texttt{NeoChip} and \texttt{Quantum Systems}, with description ``Quantum Systems owned NeoChip from 2016 until NeoChip became publicly traded."
\end{itemize}
These prompts can be tailored to the domain of the document corpus by choosing domain appropriate few-shot exemplars for in-context learning~\citep{brown2020language}.
For example, while our default prompt extracts the broad class of ``named entities'' like people, places, and organizations and is generally applicable, domains with specialized knowledge (e.g., science, medicine, law) will benefit from few-shot exemplars specialized to those domains.

The LLM can also be prompted to extract \emph{claims} about detected entities.
\emph{Claims} are important factual statements about entities, such as dates, events, and interactions with other entities.
As with entities and relationships, in-context learning exemplars can provide domain-specific guidance. Claim descriptions extracted from the example tetx chunk are as follows:
\begin{itemize}
    \item NeoChip's shares surged during their first week of trading on the NewTech Exchange.
    \item NeoChip debuted as a publicly listed company on the NewTech Exchange.
    \item Quantum Systems acquired NeoChip in 2016 and held ownership until NeoChip went public.
\end{itemize}

See \autoref{app:prompts} for prompts and details on our implementation of entity and claim extraction.

\subsubsection{Entities \& Relationships \texorpdfstring{$\rightarrow$}{-->} Knowledge Graph}
The use of an LLM to extract entities, relationships, and claims is a form of abstractive summarization -- these are meaningful summaries of concepts that, in the case of relationships and claims, may not be explicitly stated in the text.
The entity/relationship/claim extraction processes creates multiple instances of a single element because an element is typically detected and extracted multiple times across documents.

In the final step of the knowledge graph extraction process, these instances of entities and relationships become individual nodes and edges in the graph.
Entity descriptions are aggregated and summarized for each node and edge.
Relationships are aggregated into graph edges, where the number of duplicates for a given relationship becomes edge weights.
Claims are aggregated similarly.

In this manuscript, our analysis uses exact string matching for \emph{entity matching} -- the task of reconciling different extracted names for the same entity~\citep{barlaug2021neural, christen2012data, elmagarmid2006duplicate}. 
However, softer matching approaches can be used with minor adjustments to prompts or code.
Furthermore, GraphRAG is generally resilient to duplicate entities since duplicates are typically clustered together for summarization in subsequent steps.

\subsubsection{Knowledge Graph \texorpdfstring{$\rightarrow$}{-->} Graph Communities}

Given the graph index created in the previous step, a variety of community detection algorithms may be used to partition the graph into communities of strongly connected nodes (e.g., see the surveys by~\citet{fortunato2010community} and~\citet{jin2021survey}).
In our pipeline, we use Leiden community detection~\citep{Traag2019Leiden} in a hierarchical manner, recursively detecting sub-communities within each detected community until reaching leaf communities that can no longer be partitioned. 

Each level of this hierarchy provides a community partition that covers the nodes of the graph in a mutually exclusive, collectively exhaustive way, enabling divide-and-conquer global summarization.
An illustration of such hierarchical partitioning on an example dataset can be found in \autoref{app:graph}.

\subsubsection{Graph Communities \texorpdfstring{$\rightarrow$}{-->} Community Summaries}
The next step creates report-like summaries of each community in the community hierarchy, using a method designed to scale to very large datasets.
These summaries are independently useful as a way to understand the global structure and semantics of the dataset, and may themselves be used to make sense of a corpus in the absence of a specific query.
For example, a user may scan through community summaries at one level looking for general themes of interest, then read linked reports at a lower level that provide additional details for each subtopic.
Here, however, we focus on their utility as part of a graph-based index used for answering global queries.

GraphRAG generates community summaries by adding various element summaries (for nodes, edges, and related claims) to a community summary template.
Community summaries from lower-level communities are used to generate summaries for higher-level communities as follows:

\begin{itemize}
    \item \emph{Leaf-level communities}. The element summaries of a leaf-level community  are prioritized and then iteratively added to the LLM context window until the token limit is reached.
    The prioritization is as follows: for each community edge in decreasing order of combined source and target node degree (i.e., overall prominence), add descriptions of the source node, target node, the edge itself, and related claims.
    \item \emph{Higher-level communities}. If all element summaries fit within the token limit of the context window, proceed as for leaf-level communities and summarize all element summaries within the community.
    Otherwise, rank sub-communities in decreasing order of element summary tokens and iteratively substitute sub-community summaries (shorter) for their associated element summaries (longer) until they fit within the context window.
\end{itemize}

\subsubsection{Community Summaries \texorpdfstring{$\rightarrow$}{-->} Community Answers \texorpdfstring{$\rightarrow$}{-->} Global Answer}
\label{sec:querytime}

Given a user query, the community summaries generated in the previous step can be used to generate a final answer in a multi-stage process.
The hierarchical nature of the community structure also means that questions can be answered using the community summaries from different levels, raising the question of whether a particular level in the hierarchical community structure offers the best balance of summary detail and scope for general sensemaking questions (evaluated in~\autoref{sec:eval}).

For a given community level, the global answer to any user query is generated as follows:  

\begin{itemize}
    \item \emph{Prepare community summaries}. Community summaries are randomly shuffled and divided into chunks of pre-specified token size.  This ensures relevant information is distributed across chunks, rather than concentrated (and potentially lost) in a single context window.
    \item \emph{Map community answers}. Intermediate answers are generated in parallel. The LLM is also asked to generate a score between 0-100 indicating how helpful the generated answer is in answering the target question. Answers with score 0 are filtered out.
    \item \emph{Reduce to global answer}. Intermediate community answers are sorted in descending order of helpfulness score and iteratively added into a new context window until the token limit is reached. This final context is used to generate the global answer returned to the user.
\end{itemize}

\subsection{Global Sensemaking Question Generation}
To evaluate the effectiveness of RAG systems for global sensemaking tasks, we use an LLM to generate a set of corpus-specific questions designed to asses high-level understanding of a given corpus, without requiring retrieval of specific low-level facts.
Instead, given a high-level description of a corpus and its purposes, the LLM is prompted to generate personas of hypothetical users of the RAG system.
For each hypothetical user, the LLM is then prompted to specify tasks that this user would use the RAG system to complete.
Finally, for each combination of user and task, the LLM is prompted to generate questions that require understanding of the entire corpus. Algorithm 1 describes the approach.

\noindent\hrulefill

\vspace{-5px}
\textbf{Algorithm 1: Prompting Procedure for Question Generation}

\vspace{-10px}
\noindent\hrulefill

\begin{algorithmic}[1]
\State \textbf{Input:} Description of a corpus, number of users $K$, number of tasks per user $N$, number of questions per (user, task) combination $M$.
\State \textbf{Output:} A set of $K*N*M$ high-level questions requiring global understanding of the corpus.

\Procedure{GenerateQuestions}{}
    \State Based on the corpus description, prompt the LLM to:
    \begin{enumerate}
        \item Describe personas of $K$ potential users of the dataset.
        \item For each user, identify $N$ tasks relevant to the user.
        \item Specific to each user \& task pair, generate $M$ high-level questions that:
        \begin{itemize}
            \item Require understanding of the entire corpus.
            \item Do not require retrieval of specific low-level facts.
        \end{itemize}
    \end{enumerate}
    \State Collect the generated questions to produce $K*N*M$ test questions for the dataset.
\EndProcedure
\end{algorithmic}
\noindent\hrulefill
\label{alg:question_generation}

For our evaluation, we set $K=M=N=5$ for a total of 125 test questions per dataset.
\autoref{tab:sample_queries} shows example questions for each of the two evaluation datasets.
\FloatBarrier
\begin{table}
    \caption{Examples of potential users, tasks, and questions generated by the LLM based on short descriptions of the target datasets. Questions target global understanding rather than specific details.}
    \small
    \centering
    \def\arraystretch{1.25}% 
    \begin{tabular}{>{\raggedright\arraybackslash}p{0.12\textwidth}>{\raggedright\arraybackslash}p{0.88\textwidth}} \hline 
         \textbf{Dataset} & \textbf{Example activity framing and generation of global sensemaking questions}\\ \hline 
         Podcast transcripts& \emph{User}: 
A tech journalist looking for insights and trends in the tech industry
\newline
\emph{Task}: 
Understanding how tech leaders view the role of policy and regulation
\newline
\emph{Questions}:\newline
1. Which episodes deal primarily with tech policy and government regulation?
\newline
2. How do guests perceive the impact of privacy laws on technology development?
\newline
3. Do any guests discuss the balance between innovation and ethical considerations?
\newline
4. What are the suggested changes to current policies mentioned by the guests?
\newline
5. Are collaborations between tech companies and governments discussed and how?\\ \hline 
         News articles& \emph{User}: 
Educator incorporating current affairs into curricula
\newline
\emph{Task}: 
Teaching about health and wellness
\newline
\emph{Questions}:\newline
1. What current topics in health can be integrated into health education curricula?
\newline
2. How do news articles address the concepts of preventive medicine and wellness?
\newline
3. Are there examples of health articles that contradict each other, and if so, why?
\newline
 4. What insights can be gleaned about public health priorities based on news coverage?
 \newline
 5. How can educators use the dataset to highlight the importance of health literacy?\\ \hline
    \end{tabular}

    \label{tab:sample_queries}
\end{table}

\subsection{Criteria for Evaluating Global Sensemaking}
\label{sec:criteria}

Given the lack of gold standard answers to our activity-based sensemaking questions, we adopt the head-to-head comparison approach using an LLM evaluator that judges relative performance according to specific criteria.
We designed three target criteria capturing qualities that are desirable for global sensemaking activities.

\autoref{app:eval_prompts} shows the prompts for our head-to-head measures computed using an LLM evaluator, summarized as:
\begin{itemize}
    \item \emph{Comprehensiveness}. How much detail does the answer provide to cover all aspects and details of the question?
    \item \emph{Diversity}. How varied and rich is the answer in providing different perspectives and insights on the question?
    \item \emph{Empowerment}. How well does the answer help the reader understand and make informed judgments about the topic?
\end{itemize}

Furthermore, we use a ``control criterion'' called \emph{Directness} that answers \emph{``How specifically and clearly does the answer address the question?"}.
In plain terms, directness evaluates the concision of an answer in a generic sense that applies to any generated LLM summarization.
We include it to behave as a reference against which we can judge the soundness of results for the other criteria.
Since directness is effectively in opposition to comprehensiveness and diversity, we would not expect any method to win across all four criteria.

In our evaluations, the LLM is provided with the question, the generated answers from two competing systems, and prompted to compare the two answers according to the criterion before giving a final judgment of which answer is preferred. 
The LLM either indicates a winner; or, it returns a tie if they are fundamentally similar.
To account for the inherent stochasticity of LLM generation, we run each comparison with multiple replicates and average the results across replicates and questions.
An illustration of LLM assessment for answers to a sample question can be found in \autoref{app:question}.

\section{Analysis}
\label{sec:eval}

\subsection{Experiment 1}
\subsubsection{Datasets}

We selected two datasets in the one million token range, each representative of corpora that users may encounter in their real-world activities:

\textbf{Podcast transcripts}. Public transcripts of \emph{Behind the Tech with Kevin Scott}, a podcast featuring conversations between Microsoft CTO Kevin Scott and various thought leaders in science and technology ~\citep{BehindTheTech}.
This corpus was divided into 1669 $\times$ 600-token text chunks, with 100-token overlaps between chunks ($\sim$1 million tokens).

\textbf{News articles}. A benchmark dataset comprised of news articles published from September 2013 to December 2023 in a range of categories, including entertainment, business, sports, technology, health, and science ~\citep{tang2024multihop}.
The corpus is divided into 3197 $\times$ 600-token text chunks, with 100-token overlaps between chunks ($\sim$1.7 million tokens).

\subsubsection{Conditions}

We compared six conditions including GraphRAG at four different graph community levels (\textbf{C0}, \textbf{C1}, \textbf{C2}, \textbf{C3}), a text summarization method that applies our map-reduce approach directly to source texts (\textbf{TS}), and a vector RAG ``semantic search'' approach (\textbf{SS}):

\begin{itemize}
\item \textbf{CO}. Uses root-level community summaries (fewest in number) to answer user queries.
\item \textbf{C1}. Uses high-level community summaries to answer queries. These are sub-communities of C0, if present, otherwise C0 communities projected downwards.
\item \textbf{C2}. Uses intermediate-level community summaries to answer queries. These are sub-communities of C1, if present, otherwise C1 communities projected downwards.
\item \textbf{C3}. Uses low-level community summaries (greatest in number) to answer queries. These are sub-communities of C2, if present, otherwise C2 communities projected downwards.
\item \textbf{TS}. The same method as in~\autoref{sec:querytime}, except source texts (rather than community summaries) are shuffled and chunked for the map-reduce summarization stages.
\item \textbf{SS}. An implementation of vector RAG in which text chunks are retrieved and added to the available context window until the specified token limit is reached.
\end{itemize}

The size of the context window and the prompts used for answer generation are the same across all six conditions (except for minor modifications to reference styles to match the types of context information used).
Conditions only differ in how the contents of the context window are created.

The graph index supporting conditions \textbf{C0}-\textbf{C3} was created using our generic prompts for entity and relationship extraction, with entity types and few-shot examples tailored to the domain of the data. 

\subsubsection{Configuration}

We used a fixed context window size of 8k tokens for generating community summaries, community answers, and global answers (explained in \autoref{app:window}).
Graph indexing with a 600 token window (explained in \autoref{app:self_reflection}) took 281 minutes for the Podcast dataset, running on a virtual machine (16GB RAM, Intel(R) Xeon(R) Platinum 8171M CPU @ 2.60GHz) and using a public OpenAI endpoint for \texttt{gpt-4-turbo} (2M TPM, 10k RPM).

We implemented Leiden community detection using the graspologic library \citep{chung2019graspy}.
The prompts used to generate the graph index and global answers can be found in \autoref{app:sys_prompts}, while the prompts used to evaluate LLM responses against our criteria can be found in \autoref{app:eval_prompts}.
A full statistical analysis of the results presented in the next section can be found in \autoref{app:stats}.

\subsection{Experiment 2}
To validate the comprehensiveness and diversity results from Experiment 1, we implemented claim-based measures of these qualities. We use the definition of a factual claim from \citet{afacta}, which is “a statement that explicitly presents some verifiable facts.” For example, the sentence “California and New York implemented incentives for renewable energy adoption, highlighting the broader importance of sustainability in policy decisions” contains two factual claims: (1) California implemented incentives for renewable energy adoption, and (2) New York implemented incentives for renewable energy adoption.

To extract factual claims, we used \textbf{Claimify} \citep{claimify}, an LLM-based method that identifies sentences in an answer containing at least one factual claim, then decomposes these sentences into simple, self-contained factual claims. We applied Claimify to the answers generated under the conditions from Experiment 1. After removing duplicate claims from each answer, we extracted 47,075 unique claims, with an average of 31 claims per answer.

We defined two metrics, with higher values indicating better performance:
\begin{enumerate}
\item \textbf{Comprehensiveness}: Measured as the average number of claims extracted from the answers generated under each condition.
\item \textbf{Diversity}: Measured by clustering the claims for each answer and calculating the average number of clusters.
\end{enumerate}

For clustering, we followed the approach described by \citet{padmakumar}, which involved using Scikit-learn’s implementation of agglomerative clustering \citep{perdregosa}. Clusters were merged through “complete” linkage, meaning they were combined only if the maximum distance between their farthest points was less than or equal to a predefined distance threshold. The distance metric used was $1 - \text{ROUGE-L}$. Since the distance threshold influences the number of clusters, we report results across a range of thresholds.

\section{Results}

\subsection{Experiment 1}

The indexing process resulted in a graph consisting of 8,564 nodes and 20,691 edges for the Podcast dataset, and a larger graph of 15,754 nodes and 19,520 edges for the News dataset.
\autoref{tab:community summaries} shows the number of community summaries at different levels of each graph community hierarchy.

\import{}{measures_figure.tex}

\begin{table}

    \caption{Number of context units (community summaries for \textbf{C0-C3} and text chunks for \textbf{TS}), corresponding token counts, and percentage of the maximum token count. Map-reduce summarization of source texts is the most resource-intensive approach requiring the highest number of context tokens. Root-level community summaries (\textbf{C0}) require dramatically fewer tokens per query (9x-43x).}
    \small
    \centering
    \def\arraystretch{1.25}% 
    \begin{tabularx}{\textwidth}{lXXXXXXXXXX} \hline           
    & \multicolumn{5}{l}{\textbf{Podcast Transcripts}} & \multicolumn{5}{l}{\textbf{News Articles}} \\ \hline 
    & \textbf{C0} & \textbf{C1} & \textbf{C2} & \textbf{C3} & \textbf{TS} & \textbf{C0} & \textbf{C1} & \textbf{C2} & \textbf{C3} & \textbf{TS} \\ \hline 
    \textbf{Units} & 34 & 367 & 969 & 1310 & 1669 & 55 & 555 & 1797 & 2142 & 3197 \\ \hline 
    \textbf{Tokens} & 26657 & 225756 & 565720 & 746100 & 1014611 & 39770 & 352641 & 980898 & 1140266 & 1707694 \\ \hline
    \textbf{\% Max} & 2.6 & 22.2 & 55.8 & 73.5 & 100 & 2.3 & 20.7 & 57.4 & 66.8 & 100 \\ \hline
    \end{tabularx}

    \label{tab:community summaries}
\end{table}

\textbf{Global approaches vs. vector RAG}. As shown in \autoref{fig:relativemeasures} and \autoref{tab:stats_analysis}, global approaches significantly outperformed conventional vector RAG (\textbf{SS}) in both comprehensiveness and diversity criteria across datasets.
Specifically, global approaches achieved comprehensiveness win rates between 72-83\% (p\textless .001) for Podcast transcripts and 72-80\% (p\textless.001) for News articles, while diversity win rates ranged from 75-82\% (p\textless.001) and 62-71\% (p\textless.01) respectively.
Our use of directness as a validity test confirmed that vector RAG produces the most direct responses across all comparisons. 

\textbf{Empowerment}. Empowerment comparisons showed mixed results for both global approaches versus vector RAG (\textbf{SS}) and GraphRAG approaches versus source text summarization (\textbf{TS}).
Using an LLM to analyze LLM reasoning for this measure indicated that the ability to provide specific examples, quotes, and citations was judged to be key to helping users reach an informed understanding.
Tuning element extraction prompts may help to retain more of these details in the GraphRAG index.

\textbf{Community summaries vs. source texts}.  When comparing community summaries to source texts using GraphRAG, community summaries generally provided a small but consistent improvement in answer comprehensiveness and diversity, except for root-level summaries.
Intermediate-level summaries in the Podcast dataset and low-level community summaries in the News dataset achieved comprehensiveness win rates of 57\% (p\textless.001) and 64\% (p\textless.001), respectively.
Diversity win rates were 57\% (p=.036) for Podcast intermediate-level summaries and 60\% (p\textless.001) for News low-level community summaries.
\autoref{tab:community summaries} also illustrates the scalability advantages of GraphRAG compared to source text summarization: for low-level community summaries (\textbf{C3}), GraphRAG required 26-33\% fewer context tokens, while for root-level community summaries (\textbf{C0}), it required over 97\% fewer tokens. 
For a modest drop in performance compared with other global methods, root-level GraphRAG offers a highly efficient method for the iterative question answering that characterizes sensemaking activity, while retaining advantages in comprehensiveness (72\% win rate) and diversity (62\% win rate) over vector RAG.

\subsection{Experiment 2}

\autoref{tab:claim_comp} shows the results for the average number of extracted claims (i.e., the claim-based measure of comprehensiveness) per condition. For both the News and Podcast datasets, all global search conditions (\textbf{C0-C3}) and source text summarization (\textbf{TS}) had greater comprehensiveness than vector RAG (\textbf{SS}). The differences were statistically significant (p\textless.05) in all cases. These findings align with the LLM-based win rates from Experiment 1. 

\autoref{tab:claim_div} contains the results for the average number of clusters, the claim-based measure of diversity. For the Podcast dataset, all global search conditions had significantly greater diversity than \textbf{SS} across all distance thresholds (p\textless.05), consistent with the win rates observed in Experiment 1. For the News dataset, however, only \textbf{C0} significantly outperformed \textbf{SS} across all distance thresholds (p\textless.05). While \textbf{C1-C3} also achieved higher average cluster counts than \textbf{SS}, the differences were statistically significant only at certain distance thresholds. In Experiment 1, all global search conditions significantly outperformed \textbf{SS} in the News dataset – not just \textbf{C0}. However, the differences in mean diversity scores between \textbf{SS} and the global search conditions were smaller for the News dataset than for the Podcast dataset, aligning directionally with the claim-based results. 

For both comprehensiveness and diversity, across both datasets, there were no statistically significant differences observed among the global search conditions or between global search and \textbf{TS}.

Finally, for each pairwise comparison in Experiment 1, we tested whether the answer preferred by the LLM aligned with the winner based on the claim-based metrics. Since each pairwise comparison in Experiment 1 was performed five times, while the claim-based metrics provided only one outcome per comparison, we aggregated the Experiment 1 results into a single label using majority voting. For example, if \textbf{C0} won over \textbf{SS} in three out of five judgments for comprehensiveness on a given question, \textbf{C0} was labeled the winner and \textbf{SS} the loser.  However, if \textbf{C0} won twice, \textbf{SS} won once, and they tied twice, then there was no majority outcome, so the final label was a tie.

We found that exact ties were rare for the claim-based metrics. One possible solution is to define a tie based on a threshold (e.g., the absolute difference between the claim-based results for condition A and condition B must be less than or equal to $x$). However, we observed that the results were sensitive to the choice of threshold. As a result, we focused on cases where the aggregated LLM label was not a tie, representing 33\% and 39\% of pairwise comparisons for comprehensiveness and diversity, respectively. In these cases, the aggregated LLM label matched the claim-based label in 78\% of pairwise comparisons for comprehensiveness and 69-70\% for diversity (across all distance thresholds), indicating moderately strong alignment.

\begin{table}
\centering
\caption{Average number of extracted claims, reported by condition and dataset type. Bolded values represent the highest score in each column.}
\begin{tabular}{c ccc}
    \toprule
\multirow{2}{*}{\textbf{Condition}} 
        & \multicolumn{2}{c}{\textbf{Average Number of Claims}}   \\ 
    \cmidrule(lr){2-3}
        & \textbf{News Articles}  & \textbf{Podcast Transcripts}                  \\ 
    \midrule
C0    & \textbf{34.18}   & 32.21                     \\
C1    & 32.50   & 32.20                     \\
C2    & 31.62   & \textbf{32.46}                     \\
C3    & 33.14   & 32.28                     \\
TS       & 32.89   & 31.39                     \\
SS     & 25.23   & 26.50                     \\
    \bottomrule
\end{tabular}
\label{tab:claim_comp}
    \end{table}
\begin{table}
\centering
\caption{Average number of clusters across different distance thresholds, reported by condition and dataset type. Bolded values represent the highest score in each row.}
\begin{tabular}{c c ccccccc}
    \toprule
\multirow{2}{*}{\textbf{Dataset}} & \multirow{2}{*}{\textbf{Distance Threshold}} & \multicolumn{6}{c}{\textbf{Average Number of Clusters}} \\ 
    \cmidrule(lr){3-8}
 &  & \textbf{C0} & \textbf{C1} & \textbf{C2} & \textbf{C3} & \textbf{TS} & \textbf{SS} \\ 
    \midrule
\multirow{4}{*}{News Articles} & 0.5 & \textbf{23.42} & 21.85 & 21.90 & 22.13 & 21.80 & 17.92 \\
                          & 0.6 & \textbf{21.65} & 20.38 & 20.30 & 20.52 & 20.13 & 16.78 \\
                          & 0.7 & \textbf{20.19} & 19.06 & 19.03 & 19.13 & 18.62 & 15.80 \\
                          & 0.8 & \textbf{18.86} & 17.78 & 17.82 & 17.79 & 17.30 & 14.80 \\
\midrule
\multirow{4}{*}{Podcast Transcripts}  & 0.5 & \textbf{23.16} & 22.62 & 22.52 & 21.93 & 21.14 & 18.55 \\
                          & 0.6 & \textbf{21.65} & 21.33 & 21.21 & 20.62 & 19.70 & 17.39 \\
                          & 0.7 & \textbf{20.41} & 20.04 & 19.79 & 19.22 & 18.08 & 16.28 \\
                          & 0.8 & \textbf{19.26} & 18.77 & 18.46 & 17.89 & 16.66 & 15.07 \\
    \bottomrule
\end{tabular}
\label{tab:claim_div}
\end{table}

\section{Discussion}

\subsection{Limitations of evaluation approach}
Our evaluation to date has focused on sensemaking questions specific to two corpora each containing approximately 1 million tokens.
More work is needed to understand how performance generalizes to datasets from various domains with different use cases.
Comparison of fabrication rates, e.g., using approaches like SelfCheckGPT~\citep{manakul2023selfcheckgpt}, would also strengthen the current analysis.

\subsection{Future work}
The graph index, rich text annotations, and hierarchical community structure supporting the current GraphRAG approach offer many possibilities for refinement and adaptation.
This includes RAG approaches that operate in a more local manner, via embedding-based matching of user queries and graph annotations.
In particular, we see potential in hybrid RAG schemes that combine embedding-based matching with just-in-time community report generation before employing our map-reduce summarization mechanisms.
This ``roll-up'' approach could also be extended across multiple levels of the community hierarchy, as well as implemented as a more exploratory ``drill down'' mechanism that follows the information scent contained in higher-level community summaries.

\emph{Broader impacts}. As a mechanism for question answering over large document collections, there are risks to downstream sensemaking and decision-making tasks if the generated answers do not accurately represent the source data.
System use should be accompanied by clear disclosures of AI use and the potential for errors in outputs.
Compared to vector RAG, however, GraphRAG shows promise as a way to mitigate these downstream risks for questions of a global nature, which might otherwise be answered by samples of retrieved facts falsely presented as global summaries.

\section{Conclusion}

We have presented GraphRAG, a RAG approach that combines knowledge graph generation and query-focused summarization (QFS) to support human sensemaking over entire text corpora.
Initial evaluations show substantial improvements over a vector RAG baseline for both the comprehensiveness and diversity of answers, as well as favorable comparisons to a global but graph-free approach using map-reduce source text summarization. 
For situations requiring many global queries over the same dataset, summaries of root-level communities in the entity-based graph index provide a data index that is both superior to vector RAG and achieves competitive performance to other global methods at a fraction of the token cost.

\section*{Acknowledgements}
We would also like to thank the following people who contributed to the work: Alonso Guevara Fernández, Amber Hoak, Andrés Morales Esquivel, Ben Cutler, Billie Rinaldi, Chris Sanchez, Chris Trevino, Christine Caggiano, David Tittsworth, Dayenne de Souza, Douglas Orbaker, Ed Clark, Gabriel Nieves-Ponce, Gaudy Blanco Meneses, Kate Lytvynets, Katy Smith, Mónica Carvajal, Nathan Evans, Richard Ortega, Rodrigo Racanicci, Sarah Smith, and Shane Solomon. 

\medskip
%%%%%%%%%%%%%%%%%%%%%%%%%%%%%%%%%%%%%%%%%%%%%%%%%%%%%%%%%%%%
\bibliographystyle{apalike}
\bibliography{bibliography}

\begin{thebibliography}{}

\bibitem[Achiam et~al., 2023]{achiam2023gpt}
Achiam, J., Adler, S., Agarwal, S., Ahmad, L., Akkaya, I., Aleman, F.~L., Almeida, D., Altenschmidt, J., Altman, S., Anadkat, S., et~al. (2023).
\newblock Gpt-4 technical report.
\newblock {\em arXiv preprint arXiv:2303.08774}.

\bibitem[Anil et~al., 2023]{team2023gemini}
Anil, R., Borgeaud, S., Wu, Y., Alayrac, J.-B., Yu, J., Soricut, R., Schalkwyk, J., Dai, A.~M., Hauth, A., et~al. (2023).
\newblock Gemini: a family of highly capable multimodal models.
\newblock {\em arXiv preprint arXiv:2312.11805}.

\bibitem[Baek et~al., 2023]{baek2023knowledge}
Baek, J., Aji, A.~F., and Saffari, A. (2023).
\newblock Knowledge-augmented language model prompting for zero-shot knowledge graph question answering.
\newblock {\em arXiv preprint arXiv:2306.04136}.

\bibitem[Ban et~al., 2023]{ban2023query}
Ban, T., Chen, L., Wang, X., and Chen, H. (2023).
\newblock From query tools to causal architects: Harnessing large language models for advanced causal discovery from data.

\bibitem[Barlaug and Gulla, 2021]{barlaug2021neural}
Barlaug, N. and Gulla, J.~A. (2021).
\newblock Neural networks for entity matching: A survey.
\newblock {\em ACM Transactions on Knowledge Discovery from Data (TKDD)}, 15(3):1--37.

\bibitem[Baumel et~al., 2018]{baumel2018query}
Baumel, T., Eyal, M., and Elhadad, M. (2018).
\newblock Query focused abstractive summarization: Incorporating query relevance, multi-document coverage, and summary length constraints into seq2seq models.
\newblock {\em arXiv preprint arXiv:1801.07704}.

\bibitem[Blondel et~al., 2008]{blondel2008fast}
Blondel, V.~D., Guillaume, J.-L., Lambiotte, R., and Lefebvre, E. (2008).
\newblock Fast unfolding of communities in large networks.
\newblock {\em Journal of statistical mechanics: theory and experiment}, 2008(10):P10008.

\bibitem[Brown et~al., 2020]{brown2020language}
Brown, T., Mann, B., Ryder, N., Subbiah, M., Kaplan, J.~D., Dhariwal, P., Neelakantan, A., Shyam, P., Sastry, G., Askell, A., et~al. (2020).
\newblock Language models are few-shot learners.
\newblock {\em Advances in neural information processing systems}, 33:1877--1901.

\bibitem[Cheng et~al., 2024]{cheng2024lift}
Cheng, X., Luo, D., Chen, X., Liu, L., Zhao, D., and Yan, R. (2024).
\newblock Lift yourself up: Retrieval-augmented text generation with self-memory.
\newblock {\em Advances in Neural Information Processing Systems}, 36.

\bibitem[Christen and Christen, 2012]{christen2012data}
Christen, P. and Christen, P. (2012).
\newblock {\em The data matching process}.
\newblock Springer.

\bibitem[Chung et~al., 2019]{chung2019graspy}
Chung, J., Pedigo, B.~D., Bridgeford, E.~W., Varjavand, B.~K., Helm, H.~S., and Vogelstein, J.~T. (2019).
\newblock Graspy: Graph statistics in python.
\newblock {\em Journal of Machine Learning Research}, 20(158):1--7.

\bibitem[Dang, 2006]{dang2006duc}
Dang, H.~T. (2006).
\newblock Duc 2005: Evaluation of question-focused summarization systems.
\newblock In {\em Proceedings of the Workshop on Task-Focused Summarization and Question Answering}, pages 48--55.

\bibitem[Elmagarmid et~al., 2006]{elmagarmid2006duplicate}
Elmagarmid, A.~K., Ipeirotis, P.~G., and Verykios, V.~S. (2006).
\newblock Duplicate record detection: A survey.
\newblock {\em IEEE Transactions on knowledge and data engineering}, 19(1):1--16.

\bibitem[Es et~al., 2023]{es2023ragas}
Es, S., James, J., Espinosa-Anke, L., and Schockaert, S. (2023).
\newblock Ragas: Automated evaluation of retrieval augmented generation.
\newblock {\em arXiv preprint arXiv:2309.15217}.

\bibitem[Etzioni et~al., 2004]{knowitall}
Etzioni, O., Cafarella, M., Downey, D., Kok, S., Popescu, A.-M., Shaked, T., Soderland, S., Weld, D.~S., and Yates, A. (2004).
\newblock Web-scale information extraction in knowitall: (preliminary results).
\newblock In {\em Proceedings of the 13th International Conference on World Wide Web}, WWW '04, page 100–110, New York, NY, USA. Association for Computing Machinery.

\bibitem[Feng et~al., 2023]{feng2023retrieval}
Feng, Z., Feng, X., Zhao, D., Yang, M., and Qin, B. (2023).
\newblock Retrieval-generation synergy augmented large language models.
\newblock {\em arXiv preprint arXiv:2310.05149}.

\bibitem[Fortunato, 2010]{fortunato2010community}
Fortunato, S. (2010).
\newblock Community detection in graphs.
\newblock {\em Physics reports}, 486(3-5):75--174.

\bibitem[Gao et~al., 2023]{gao2023retrieval}
Gao, Y., Xiong, Y., Gao, X., Jia, K., Pan, J., Bi, Y., Dai, Y., Sun, J., and Wang, H. (2023).
\newblock Retrieval-augmented generation for large language models: A survey.
\newblock {\em arXiv preprint arXiv:2312.10997}.

\bibitem[He et~al., 2024]{he2024g}
He, X., Tian, Y., Sun, Y., Chawla, N.~V., Laurent, T., LeCun, Y., Bresson, X., and Hooi, B. (2024).
\newblock G-retriever: Retrieval-augmented generation for textual graph understanding and question answering.
\newblock {\em arXiv preprint arXiv:2402.07630}.

\bibitem[Huang et~al., 2023]{huang2023large}
Huang, J., Chen, X., Mishra, S., Zheng, H.~S., Yu, A.~W., Song, X., and Zhou, D. (2023).
\newblock Large language models cannot self-correct reasoning yet.
\newblock {\em arXiv preprint arXiv:2310.01798}.

\bibitem[Jacomy et~al., 2014]{fa2}
Jacomy, M., Venturini, T., Heymann, S., and Bastian, M. (2014).
\newblock Forceatlas2, a continuous graph layout algorithm for handy network visualization designed for the gephi software.
\newblock {\em PLoS ONE 9(6): e98679. https://doi.org/10.1371/journal.pone.0098679}.

\bibitem[Jin et~al., 2021]{jin2021survey}
Jin, D., Yu, Z., Jiao, P., Pan, S., He, D., Wu, J., Philip, S.~Y., and Zhang, W. (2021).
\newblock A survey of community detection approaches: From statistical modeling to deep learning.
\newblock {\em IEEE Transactions on Knowledge and Data Engineering}, 35(2):1149--1170.

\bibitem[Kang et~al., 2023]{kang2023knowledge}
Kang, M., Kwak, J.~M., Baek, J., and Hwang, S.~J. (2023).
\newblock Knowledge graph-augmented language models for knowledge-grounded dialogue generation.
\newblock {\em arXiv preprint arXiv:2305.18846}.

\bibitem[Khattab et~al., 2022]{khattab2022demonstrate}
Khattab, O., Santhanam, K., Li, X.~L., Hall, D., Liang, P., Potts, C., and Zaharia, M. (2022).
\newblock Demonstrate-search-predict: Composing retrieval and language models for knowledge-intensive nlp.
\newblock {\em arXiv preprint arXiv:2212.14024}.

\bibitem[Kim et~al., 2016]{kimprob}
Kim, D., Xie, L., and Ong, C.~S. (2016).
\newblock Probabilistic knowledge graph construction: Compositional and incremental approaches.
\newblock In {\em Proceedings of the 25th ACM International on Conference on Information and Knowledge Management}, CIKM '16, page 2257–2262, New York, NY, USA. Association for Computing Machinery.

\bibitem[Kim et~al., 2023]{kim2023tree}
Kim, G., Kim, S., Jeon, B., Park, J., and Kang, J. (2023).
\newblock Tree of clarifications: Answering ambiguous questions with retrieval-augmented large language models.
\newblock {\em arXiv preprint arXiv:2310.14696}.

\bibitem[Klein et~al., 2006]{klein2006amaking}
Klein, G., Moon, B., and Hoffman, R.~R. (2006).
\newblock Making sense of sensemaking 1: Alternative perspectives.
\newblock {\em IEEE intelligent systems}, 21(4):70--73.

\bibitem[Kosinski, 2024]{kosinski2024evaluating}
Kosinski, M. (2024).
\newblock Evaluating large language models in theory of mind tasks.
\newblock {\em Proceedings of the National Academy of Sciences}, 121(45):e2405460121.

\bibitem[Kuratov et~al., 2024]{kuratov2024search}
Kuratov, Y., Bulatov, A., Anokhin, P., Sorokin, D., Sorokin, A., and Burtsev, M. (2024).
\newblock In search of needles in a 11m haystack: Recurrent memory finds what llms miss.

\bibitem[LangChain, 2024]{LangchainGraphRAG}
LangChain (2024).
\newblock Langchain graphs.
\newblock \url{https://langchain-graphrag.readthedocs.io/en/latest/}.

\bibitem[Laskar et~al., 2020]{laskar2020query}
Laskar, M. T.~R., Hoque, E., and Huang, J. (2020).
\newblock Query focused abstractive summarization via incorporating query relevance and transfer learning with transformer models.
\newblock In {\em Advances in Artificial Intelligence: 33rd Canadian Conference on Artificial Intelligence, Canadian AI 2020, Ottawa, ON, Canada, May 13--15, 2020, Proceedings 33}, pages 342--348. Springer.

\bibitem[Lewis et~al., 2020]{lewis2020retrieval}
Lewis, P., Perez, E., Piktus, A., Petroni, F., Karpukhin, V., Goyal, N., K{\"u}ttler, H., Lewis, M., Yih, W.-t., Rockt{\"a}schel, T., et~al. (2020).
\newblock Retrieval-augmented generation for knowledge-intensive nlp tasks.
\newblock {\em Advances in Neural Information Processing Systems}, 33:9459--9474.

\bibitem[Liu et~al., 2023]{liu-etal:2023:tacl}
Liu, N.~F., Lin, K., Hewitt, J., Paranjape, A., Bevilacqua, M., Petroni, F., and Liang, P. (2023).
\newblock Lost in the middle: How language models use long contexts.
\newblock arXiv:2307.03172.

\bibitem[{LlamaIndex}, 2024]{LlamaIndexGraphRAG}
{LlamaIndex} (2024).
\newblock {GraphRAG Implementation with LlamaIndex - V2}.
\newblock \url{https://github.com/run-llama/llama_index/blob/main/docs/docs/examples/cookbooks/GraphRAG_v2.ipynb}.

\bibitem[Madaan et~al., 2024]{madaan2024self}
Madaan, A., Tandon, N., Gupta, P., Hallinan, S., Gao, L., Wiegreffe, S., Alon, U., Dziri, N., Prabhumoye, S., Yang, Y., et~al. (2024).
\newblock Self-refine: Iterative refinement with self-feedback.
\newblock {\em Advances in Neural Information Processing Systems}, 36.

\bibitem[Manakul et~al., 2023]{manakul2023selfcheckgpt}
Manakul, P., Liusie, A., and Gales, M.~J. (2023).
\newblock Selfcheckgpt: Zero-resource black-box hallucination detection for generative large language models.
\newblock {\em arXiv preprint arXiv:2303.08896}.

\bibitem[Mao et~al., 2020]{mao2020generation}
Mao, Y., He, P., Liu, X., Shen, Y., Gao, J., Han, J., and Chen, W. (2020).
\newblock Generation-augmented retrieval for open-domain question answering.
\newblock {\em arXiv preprint arXiv:2009.08553}.

\bibitem[Martin et~al., 2011]{martinbrown}
Martin, S., Brown, W.~M., Klavans, R., and Boyack, K. (2011).
\newblock Openord: An open-source toolbox for large graph layout.
\newblock {\em SPIE Conference on Visualization and Data Analysis (VDA)}.

\bibitem[Melnyk et~al., 2022]{melnyk2022knowledge}
Melnyk, I., Dognin, P., and Das, P. (2022).
\newblock Knowledge graph generation from text.

\bibitem[Metropolitansky and Larson, 2025]{claimify}
Metropolitansky, D. and Larson, J. (2025).
\newblock Towards effective extraction and evaluation of factual claims.

\bibitem[{{Microsoft}}, 2023]{ai4science2023impact}
{{Microsoft}} (2023).
\newblock The impact of large language models on scientific discovery: a preliminary study using gpt-4.

\bibitem[Mooney and Bunescu, 2005]{mooney}
Mooney, R.~J. and Bunescu, R. (2005).
\newblock Mining knowledge from text using information extraction.
\newblock {\em SIGKDD Explor. Newsl.}, 7(1):3–10.

\bibitem[NebulaGraph, 2024]{NebulaGraph}
NebulaGraph (2024).
\newblock Nebulagraph launches industry-first graph rag: Retrieval-augmented generation with llm based on knowledge graphs.
\newblock \url{https://www.nebula-graph.io/posts/graph-RAG}.

\bibitem[Neo4J, 2024]{Neo4jGraphRAG}
Neo4J (2024).
\newblock Get started with graphrag: Neo4j’s ecosystem tools.
\newblock \url{https://neo4j.com/developer-blog/graphrag-ecosystem-tools/}.

\bibitem[Newman, 2006]{newman2006modularity}
Newman, M.~E. (2006).
\newblock Modularity and community structure in networks.
\newblock {\em Proceedings of the national academy of sciences}, 103(23):8577--8582.

\bibitem[Ni et~al., 2024]{afacta}
Ni, J., Shi, M., Stammbach, D., Sachan, M., Ash, E., and Leippold, M. (2024).
\newblock {AF}a{CTA}: Assisting the annotation of factual claim detection with reliable {LLM} annotators.
\newblock In Ku, L.-W., Martins, A., and Srikumar, V., editors, {\em Proceedings of the 62nd Annual Meeting of the Association for Computational Linguistics (Volume 1: Long Papers)}, pages 1890--1912, Bangkok, Thailand. Association for Computational Linguistics.

\bibitem[OpenAI, 2023]{openai2023chatgpt}
OpenAI (2023).
\newblock Chatgpt: Gpt-4 language model.

\bibitem[Padmakumar and He, 2024]{padmakumar}
Padmakumar, V. and He, H. (2024).
\newblock Does writing with language models reduce content diversity?
\newblock {\em ICLR}.

\bibitem[Pedregosa et~al., 2011]{perdregosa}
Pedregosa, F., Varoquaux, G., Gramfort, A., Michel, V., Thirion, B., Grisel, O., Blondel, M., Prettenhofer, P., Weiss, R., Dubourg, V., Vanderplas, J., Passos, A., Cournapeau, D., Brucher, M., Perrot, M., and Duchesnay, E. (2011).
\newblock Scikit-learn: Machine learning in python.
\newblock {\em Journal of Machine Learning Research}, 12:2825--2830.

\bibitem[Ram et~al., 2023]{ram2023context}
Ram, O., Levine, Y., Dalmedigos, I., Muhlgay, D., Shashua, A., Leyton-Brown, K., and Shoham, Y. (2023).
\newblock In-context retrieval-augmented language models.
\newblock {\em Transactions of the Association for Computational Linguistics}, 11:1316--1331.

\bibitem[Ranade and Joshi, 2023]{ranade2023fabula}
Ranade, P. and Joshi, A. (2023).
\newblock Fabula: Intelligence report generation using retrieval-augmented narrative construction.
\newblock {\em arXiv preprint arXiv:2310.13848}.

\bibitem[Salminen et~al., 2024]{salminen2024deus}
Salminen, J., Liu, C., Pian, W., Chi, J., H{\"a}yh{\"a}nen, E., and Jansen, B.~J. (2024).
\newblock Deus ex machina and personas from large language models: Investigating the composition of ai-generated persona descriptions.
\newblock In {\em Proceedings of the CHI Conference on Human Factors in Computing Systems}, pages 1--20.

\bibitem[Sarthi et~al., 2024]{sarthi2024raptor}
Sarthi, P., Abdullah, S., Tuli, A., Khanna, S., Goldie, A., and Manning, C.~D. (2024).
\newblock Raptor: Recursive abstractive processing for tree-organized retrieval.
\newblock {\em arXiv preprint arXiv:2401.18059}.

\bibitem[Scott, 2024]{BehindTheTech}
Scott, K. (2024).
\newblock {Behind the Tech}.
\newblock \url{https://www.microsoft.com/en-us/behind-the-tech}.

\bibitem[Shao et~al., 2023]{shao2023enhancing}
Shao, Z., Gong, Y., Shen, Y., Huang, M., Duan, N., and Chen, W. (2023).
\newblock Enhancing retrieval-augmented large language models with iterative retrieval-generation synergy.
\newblock {\em arXiv preprint arXiv:2305.15294}.

\bibitem[Shin et~al., 2024]{shin2024understanding}
Shin, J., Hedderich, M.~A., Rey, B.~J., Lucero, A., and Oulasvirta, A. (2024).
\newblock Understanding human-ai workflows for generating personas.
\newblock In {\em Proceedings of the 2024 ACM Designing Interactive Systems Conference}, pages 757--781.

\bibitem[Shinn et~al., 2024]{shinn2024reflexion}
Shinn, N., Cassano, F., Gopinath, A., Narasimhan, K., and Yao, S. (2024).
\newblock Reflexion: Language agents with verbal reinforcement learning.
\newblock {\em Advances in Neural Information Processing Systems}, 36.

\bibitem[Su et~al., 2020]{su2020caire}
Su, D., Xu, Y., Yu, T., Siddique, F.~B., Barezi, E.~J., and Fung, P. (2020).
\newblock Caire-covid: A question answering and query-focused multi-document summarization system for covid-19 scholarly information management.
\newblock {\em arXiv preprint arXiv:2005.03975}.

\bibitem[Tan et~al., 2017]{representationlearning}
Tan, Z., Zhao, X., and Wang, W. (2017).
\newblock Representation learning of large-scale knowledge graphs via entity feature combinations.
\newblock In {\em Proceedings of the 2017 ACM on Conference on Information and Knowledge Management}, CIKM '17, page 1777–1786, New York, NY, USA. Association for Computing Machinery.

\bibitem[Tang and Yang, 2024]{tang2024multihop}
Tang, Y. and Yang, Y. (2024).
\newblock {MultiHop-RAG}: Benchmarking retrieval-augmented generation for multi-hop queries.
\newblock {\em arXiv preprint arXiv:2401.15391}.

\bibitem[Touvron et~al., 2023]{touvron2023llama}
Touvron, H., Martin, L., Stone, K., Albert, P., Almahairi, A., Babaei, Y., Bashlykov, N., Batra, S., Bhargava, P., Bhosale, S., et~al. (2023).
\newblock Llama 2: Open foundation and fine-tuned chat models.
\newblock {\em arXiv preprint arXiv:2307.09288}.

\bibitem[Traag et~al., 2019]{Traag2019Leiden}
Traag, V.~A., Waltman, L., and Van~Eck, N.~J. (2019).
\newblock From {L}ouvain to {L}eiden: guaranteeing well-connected communities.
\newblock {\em Scientific Reports}, 9(1).

\bibitem[Trajanoska et~al., 2023]{Trajanoska2023EnhancingKG}
Trajanoska, M., Stojanov, R., and Trajanov, D. (2023).
\newblock Enhancing knowledge graph construction using large language models.
\newblock {\em ArXiv}, abs/2305.04676.

\bibitem[Trivedi et~al., 2022]{trivedi2022interleaving}
Trivedi, H., Balasubramanian, N., Khot, T., and Sabharwal, A. (2022).
\newblock Interleaving retrieval with chain-of-thought reasoning for knowledge-intensive multi-step questions.
\newblock {\em arXiv preprint arXiv:2212.10509}.

\bibitem[Wang et~al., 2023a]{wang2023chatgpt}
Wang, J., Liang, Y., Meng, F., Sun, Z., Shi, H., Li, Z., Xu, J., Qu, J., and Zhou, J. (2023a).
\newblock Is chatgpt a good nlg evaluator? a preliminary study.
\newblock {\em arXiv preprint arXiv:2303.04048}.

\bibitem[Wang et~al., 2024]{wang2024feb4rag}
Wang, S., Khramtsova, E., Zhuang, S., and Zuccon, G. (2024).
\newblock Feb4rag: Evaluating federated search in the context of retrieval augmented generation.
\newblock {\em arXiv preprint arXiv:2402.11891}.

\bibitem[Wang et~al., 2022]{wang2022self}
Wang, X., Wei, J., Schuurmans, D., Le, Q., Chi, E., Narang, S., Chowdhery, A., and Zhou, D. (2022).
\newblock Self-consistency improves chain of thought reasoning in language models.
\newblock {\em arXiv preprint arXiv:2203.11171}.

\bibitem[Wang et~al., 2023b]{wang2023knowledge}
Wang, Y., Lipka, N., Rossi, R.~A., Siu, A., Zhang, R., and Derr, T. (2023b).
\newblock Knowledge graph prompting for multi-document question answering.

\bibitem[Xu and Lapata, 2021]{xu2021text}
Xu, Y. and Lapata, M. (2021).
\newblock Text summarization with latent queries.
\newblock {\em arXiv preprint arXiv:2106.00104}.

\bibitem[Yang et~al., 2018]{yang2018hotpotqa}
Yang, Z., Qi, P., Zhang, S., Bengio, Y., Cohen, W.~W., Salakhutdinov, R., and Manning, C.~D. (2018).
\newblock {HotpotQA}: A dataset for diverse, explainable multi-hop question answering.
\newblock In {\em Conference on Empirical Methods in Natural Language Processing ({EMNLP})}.

\bibitem[Yao et~al., 2017]{yao2017recent}
Yao, J.-g., Wan, X., and Xiao, J. (2017).
\newblock Recent advances in document summarization.
\newblock {\em Knowledge and Information Systems}, 53:297--336.

\bibitem[Yao et~al., 2023]{yao2023exploring}
Yao, L., Peng, J., Mao, C., and Luo, Y. (2023).
\newblock Exploring large language models for knowledge graph completion.

\bibitem[Yates et~al., 2007]{yates-etal-2007-textrunner}
Yates, A., Banko, M., Broadhead, M., Cafarella, M., Etzioni, O., and Soderland, S. (2007).
\newblock {T}ext{R}unner: Open information extraction on the web.
\newblock In Carpenter, B., Stent, A., and Williams, J.~D., editors, {\em Proceedings of Human Language Technologies: The Annual Conference of the North {A}merican Chapter of the Association for Computational Linguistics ({NAACL}-{HLT})}, pages 25--26, Rochester, New York, USA. Association for Computational Linguistics.

\bibitem[Yuan et~al., 2024]{yuan2024s}
Yuan, X., Li, J., Wang, D., Chen, Y., Mao, X., Huang, L., Xue, H., Wang, W., Ren, K., and Wang, J. (2024).
\newblock S-eval: Automatic and adaptive test generation for benchmarking safety evaluation of large language models.
\newblock {\em arXiv preprint arXiv:2405.14191}.

\bibitem[Zhang, 2023]{zhang2023graph}
Zhang, J. (2023).
\newblock Graph-toolformer: To empower llms with graph reasoning ability via prompt augmented by chatgpt.
\newblock {\em arXiv preprint arXiv:2304.11116}.

\bibitem[Zhang et~al., 2024a]{zhang2024causal}
Zhang, Y., Zhang, Y., Gan, Y., Yao, L., and Wang, C. (2024a).
\newblock Causal graph discovery with retrieval-augmented generation based large language models.
\newblock {\em arXiv preprint arXiv:2402.15301}.

\bibitem[Zhang et~al., 2024b]{zhang2024darg}
Zhang, Z., Chen, J., and Yang, D. (2024b).
\newblock Darg: Dynamic evaluation of large language models via adaptive reasoning graph.
\newblock {\em arXiv preprint arXiv:2406.17271}.

\bibitem[Zheng et~al., 2024]{zheng2024judging}
Zheng, L., Chiang, W.-L., Sheng, Y., Zhuang, S., Wu, Z., Zhuang, Y., Lin, Z., Li, Z., Li, D., Xing, E., et~al. (2024).
\newblock Judging llm-as-a-judge with mt-bench and chatbot arena.
\newblock {\em Advances in Neural Information Processing Systems}, 36.

\bibitem[Zhu et~al., 2024]{zhu2024llms}
Zhu, Y., Wang, X., Chen, J., Qiao, S., Ou, Y., Yao, Y., Deng, S., Chen, H., and Zhang, N. (2024).
\newblock Llms for knowledge graph construction and reasoning: Recent capabilities and future opportunities.

\end{thebibliography}

\appendix

\newpage
\pagebreak
\section{Entity and Relationship Extraction Approach}
\label{app:prompts}

The following prompts, designed for GPT-4, are used in the default GraphRAG initialization pipeline:
\begin{itemize}
\item \href{https://github.com/microsoft/graphrag/blob/6d21ef268377e319a165ca2250bd6841737df1ad/graphrag/prompts/index/entity_extraction.py#L6}{Default Graph Extraction Prompt}
\item \href{https://github.com/microsoft/graphrag/blob/6d21ef268377e319a165ca2250bd6841737df1ad/graphrag/prompts/index/claim_extraction.py#L1}{Claim Extraction Prompt}
\end{itemize}

\subsection{Entity Extraction}
\label{app:entity}
We do this using a multipart LLM prompt that first identifies all \emph{entities} in the text, including their name, type, and description, before identifying all \emph{relationships} between clearly related entities, including the source and target entities and a description of their relationship. Both kinds of element instance are output in a single list of delimited tuples.

\subsection{Self-Reflection}
\label{app:self_reflection}

The choice of prompt engineering techniques has a strong impact on the quality of knowledge graph extraction ~\citep{zhu2024llms}, and different techniques have different costs in terms of tokens consumed and generated by the model.
\emph{Self-reflection} is a prompt engineering technique where the LLM generates an answer, and is then prompted to evaluate its output for correctness, clarity, or completeness, then finally generate an improved response based on that evaluation \citep{huang2023large, shinn2024reflexion, wang2022self, madaan2024self}.
We leverage self-reflection in knowledge graph extraction, and explore ways how removing self-reflection affects performance and cost.

Using larger chunk size is less costly in terms of calls to the LLM.
However, the LLM tends to extract few entities from chunks of larger size.
For example, in a sample dataset (HotPotQA,~\citeauthor{yang2018hotpotqa},~\citeyear{yang2018hotpotqa}), GPT-4 extracted almost twice as many entity references when the chunk size was 600 tokens than when it was 2400.
To address this issue, we deploy a self-reflection prompt engineering approach.
After entities are extracted from a chunk, we provide the extracted entities back to the LLM, prompting it to ``glean'' any entities that it may have missed.
This is a multi-stage process in which we first ask the LLM to assess whether all entities were extracted, using a logit bias of 100 to force a yes/no decision.
If the LLM responds that entities were missed, then a continuation indicating that ``MANY entities were missed in the last extraction'' encourages the LLM to detect these missing entities.
This approach allows us to use larger chunk sizes without a drop in quality (\autoref{fig:chunkentities}) or the forced introduction of noise.
We interate self-reflection steps up to a specified maximum number of times.

\begin{figure}[h]
    \centering
    \begin{tikzpicture}
        \begin{axis}[
            xlabel={Number of self-reflection iterations performed},
            ylabel={Entity references detected},
            ytick={0,10000,20000,30000},
            yticklabels={0,10000,20000,30000},
            ymin=0,
            ymax=30000,
            xtick={0,1,2,3},
            legend pos=north west,
            grid=both,
            width=14cm,
            height=5.1cm,
            scaled y ticks=false
        ]
        
        \addplot[mark=o, blue!70] coordinates {
            (0, 9348)
            (1, 15976)
            (2, 19491)
            (3, 27240)
        };
        \addlegendentry{600 chunk size};

        \addplot[mark=o, ForestGreen] coordinates {
            (0, 7119)
            (1, 12877)
            (2, 17794)
            (3, 22399)
        };
        \addlegendentry{1200 chunk size};

        \addplot[mark=o, orange!80] coordinates {
            (0, 5761)
            (1, 10606)
            (2, 14897)
            (3, 19433)
        };
        \addlegendentry{2400 chunk size};
        \end{axis}

    \end{tikzpicture}
    \caption{How the entity references detected in the HotPotQA dataset~\citep{yang2018hotpotqa}}  varies with chunk size and self-reflection iterations for our generic entity extraction prompt with \texttt{gpt-4-turbo}.
    \label{fig:chunkentities}
\end{figure}

\section{Example Community Detection}
\label{app:graph}

\begin{figure}[ht]
    \centering
    \begin{tabular}{cc}
        \includegraphics[width=0.5\textwidth]{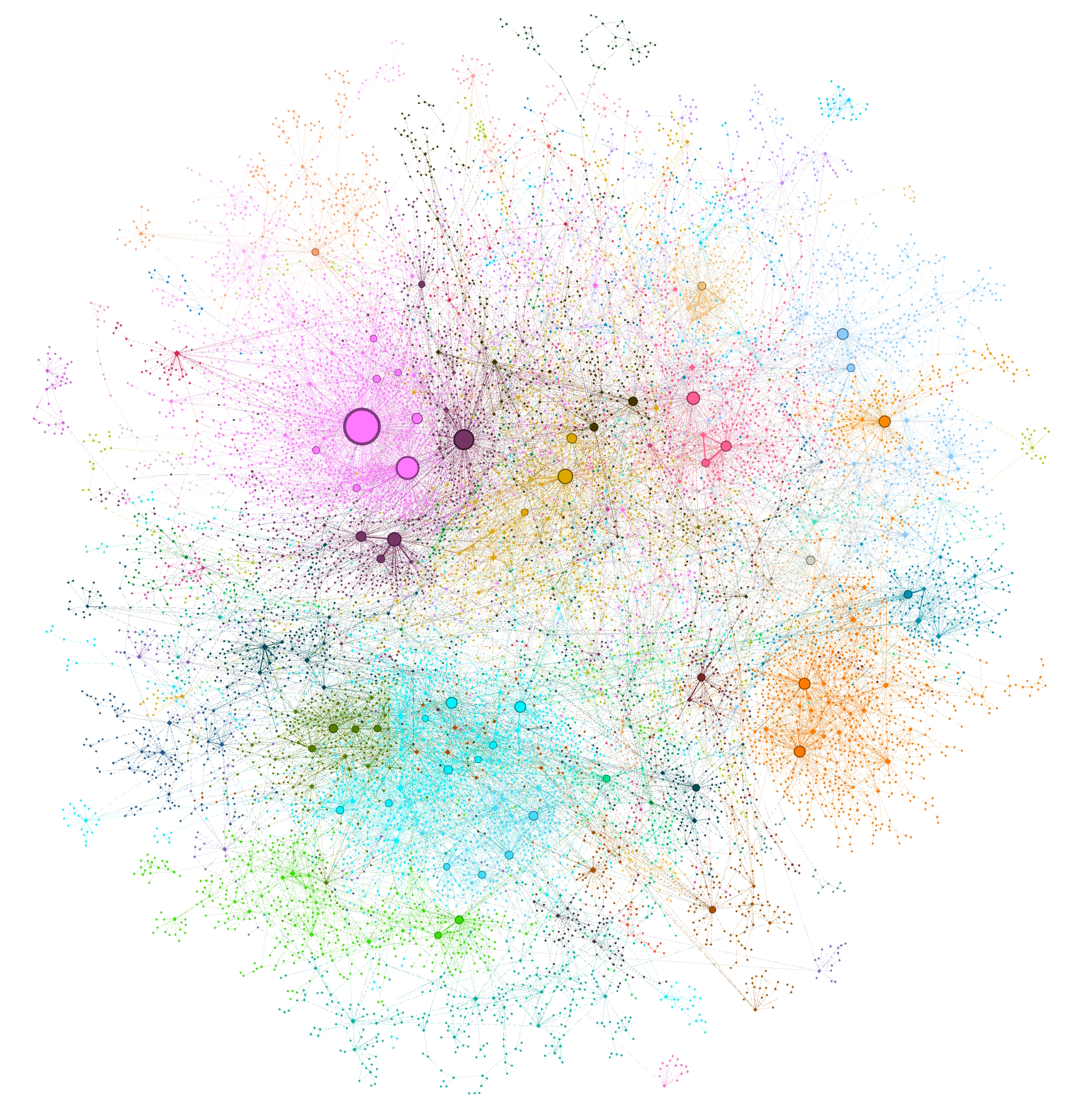}&
        \includegraphics[width=0.5\textwidth]{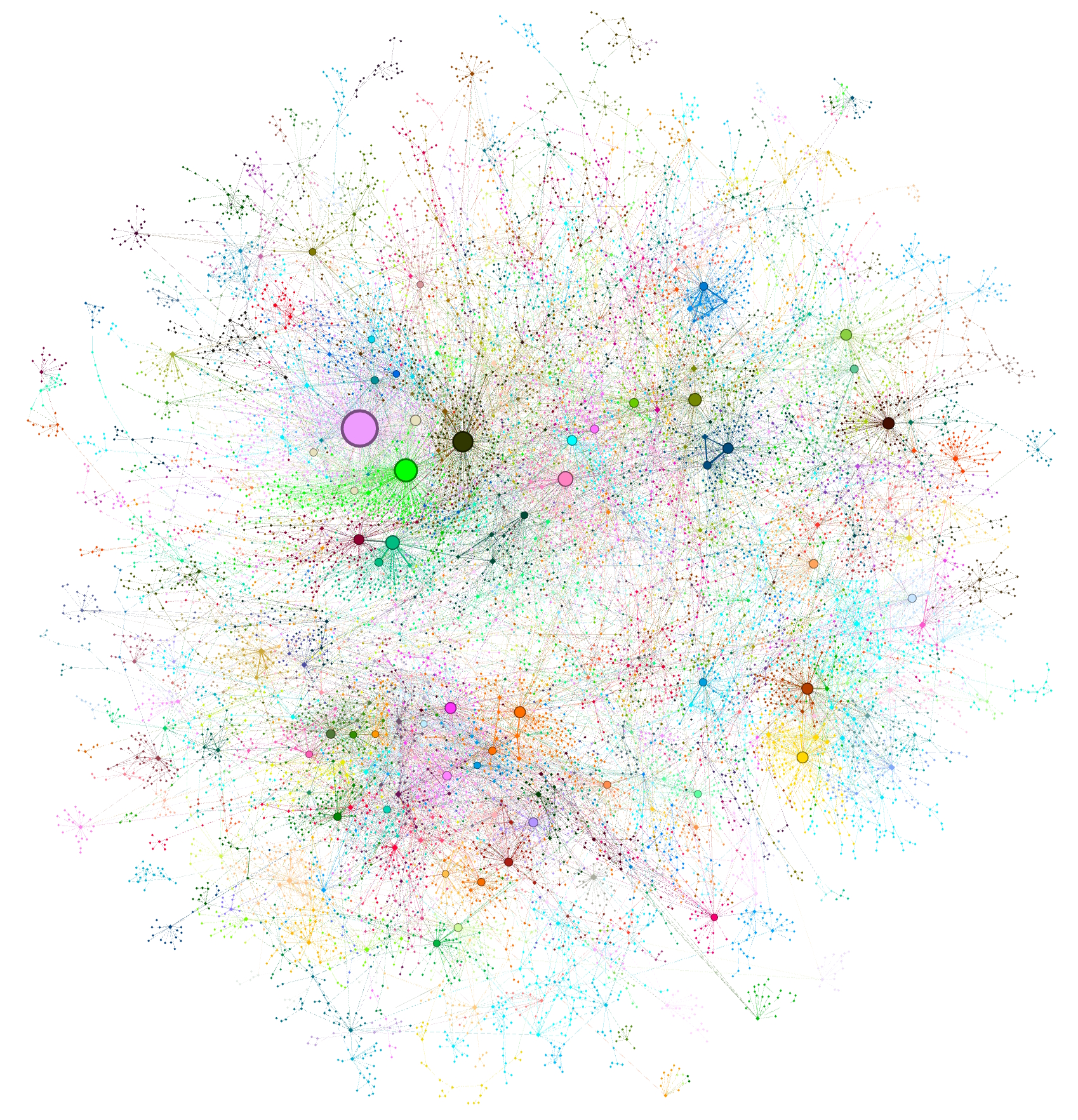}\\
        (a) Root communities at level 0 & (b) Sub-communities at level 1
    \end{tabular}
    \caption{Graph communities detected using the Leiden algorithm~\citep{Traag2019Leiden} over the MultiHop-RAG~\citep{tang2024multihop} dataset as indexed. Circles represent entity nodes with size proportional to their degree. Node layout was performed via OpenORD~\citep{martinbrown}  and Force Atlas 2~\citep{fa2}. Node colors represent entity communities, shown at two levels of hierarchical clustering: (a) Level 0, corresponding to the hierarchical partition with maximum modularity, and (b) Level 1, which reveals internal structure within these root-level communities.}
    \label{fig:graphlayout}
\end{figure}

\raggedbottom

\pagebreak
\section{Context Window Selection}
\label{app:window}
The effect of context window size on any particular task is unclear, especially for models like \texttt{gpt-4-turbo} with a large context size of 128k tokens. Given the potential for information to be ``lost in the middle'' of longer contexts~\citep{liu-etal:2023:tacl, kuratov2024search}, we wanted to explore the effects of varying the context window size for our combinations of datasets, questions, and metrics. In particular, our goal was to determine the optimum context size for our baseline condition (\textbf{SS}) and then use this uniformly for all query-time LLM use. To that end, we tested four context window sizes: 8k, 16k, 32k and 64k. Surprisingly, the smallest context window size tested (8k) was universally better for all comparisons on comprehensiveness (average win rate of 58.1\%), while performing comparably with larger context sizes on diversity (average win rate = 52.4\%), and empowerment (average win rate = 51.3\%). Given our preference for more comprehensive and diverse answers, we therefore used a fixed context window size of 8k tokens for the final evaluation.

\raggedbottom

\pagebreak
\section{Example Answer Comparison}
\label{app:question}
\begin{table}[h]
    \caption{Example question, answers, and LLM-generated assessments for the News article dataset.}
    \small
    \centering
    \def\arraystretch{1.25}% 
    \begin{tabular}{>{\raggedright\arraybackslash}p{0.08\textwidth}>{\raggedright\arraybackslash}p{0.92\textwidth}} \hline 
 \textbf{Question} & \textbf{Which public figures are repeatedly mentioned across various entertainment articles?}\\ \hline 
 \textbf{Graph RAG} & The entertainment industry is vast and diverse, encompassing film, television, music, sports, and digital media. Certain public figures stand out due to their significant contributions and influence across these sectors. The following summary highlights key individuals who are repeatedly mentioned in various entertainment articles, reflecting their impact and presence within the industry.

\textbf{Actors and Directors} [...] \textbf{Public Figures in Controversy} [...]
 \textbf{Musicians and Executives} [...] \textbf{Athletes and Coaches} [...] \textbf{Influencers and Entrepreneurs} [...]

The repeated mention of these figures in entertainment articles signifies their ongoing relevance and the public's interest in their work. Their influence spans across various aspects of entertainment, from shaping cultural narratives in film and television to driving trends in music and digital media. These individuals not only contribute to their respective fields but also influence the broader cultural landscape, often becoming central figures in social discussions and public discourse.\vspace*{2mm}\\ \hline 
\textbf{Na\"ive RAG} & Public figures who are repeatedly mentioned across various entertainment articles include Taylor Swift, Travis Kelce, Britney Spears, and Justin Timberlake. These individuals have been highlighted for various reasons, ranging from their professional achievements to their personal lives.

\textbf{Taylor Swift} [...] \textbf{Travis Kelce} [...] \textbf{Britney Spears} [...] \textbf{Justin Timberlake}
[...]

These figures are frequently covered due to their high-profile status and the public's interest in their careers and personal lives. Their activities, whether in music, sports, or personal relationships, have significant cultural and economic impacts, as evidenced by the media coverage and public reactions.
\vspace*{2mm}\\ \hline 
\textbf{LLM \break Decision} & \textbf{Comprehensiveness: Winner=1 (Graph RAG)}. Answer 1 is better because it provides a more comprehensive and detailed list of public figures from a wider range of entertainment sectors, including film, television, music, sports, gaming, and digital media. It also includes examples of their contributions and the impact they have on their respective fields, as well as mentions of controversies and their implications. Answer 2, while detailed in its coverage of a few individuals, is limited to a smaller number of public figures and focuses primarily on their personal lives and relationships rather than a broad spectrum of their professional influence across the entertainment industry.
\vspace*{2mm}
\newline
\textbf{Diversity: Winner=1 (Graph RAG)}. Answer 1 is better because it provides a more varied and rich response by covering a wide range of public figures from different sectors of the entertainment industry, including film, television, music, sports, gaming, and digital media. It offers insights into the contributions and influence of these figures, as well as controversies and their impact on public discourse. The answer also cites specific data sources for each mentioned figure, indicating a diverse range of evidence to support the claims. In contrast, Answer 2 focuses on a smaller group of public figures, primarily from the music industry and sports, and relies heavily on a single source for data, which makes it less diverse in perspectives and insights.
\vspace*{2mm}
\newline
\textbf{Empowerment: Winner=1 (Graph RAG)}. Answer 1 is better because it provides a comprehensive and structured overview of public figures across various sectors of the entertainment industry, including film, television, music, sports, and digital media. It lists multiple individuals, providing specific examples of their contributions and the context in which they are mentioned in entertainment articles, along with references to data reports for each claim. This approach helps the reader understand the breadth of the topic and make informed judgments without being misled. In contrast, Answer 2 focuses on a smaller group of public figures and primarily discusses their personal lives and relationships, which may not provide as broad an understanding of the topic. While Answer 2 also cites sources, it does not match the depth and variety of Answer 1.
\vspace*{2mm}
\newline
\textbf{Directness: Winner=2 (Na\"ive RAG)}. Answer 2 is better because it directly lists specific public figures who are repeatedly mentioned across various entertainment articles, such as Taylor Swift, Travis Kelce, Britney Spears, and Justin Timberlake, and provides concise explanations for their frequent mentions. Answer 1, while comprehensive, includes a lot of detailed information about various figures in different sectors of entertainment, which, while informative, does not directly answer the question with the same level of conciseness and specificity as Answer 2.\vspace*{2mm}\\ \hline
    \end{tabular}
    \label{tab:example_assessment}
\end{table}

\raggedbottom

\pagebreak
\section{System Prompts}
\label{app:sys_prompts}

\subsection{Element Instance Generation}

\scriptsize
\texttt{---Goal---\\
Given a text document that is potentially relevant to this activity and a list of entity types, identify all entities of those types from the text and all relationships among the identified entities.\\
\\
---Steps---\\
1. Identify all entities. For each identified entity, extract the following information:\\
- entity\_name: Name of the entity, capitalized\\
- entity\_type: One of the following types: [\{entity\_types\}]\\
- entity\_description: Comprehensive description of the entity's attributes and activities\\
\\
Format each entity as ("entity"\{tuple\_delimiter\}<entity\_name>\{tuple\_delimiter\}<entity\_type>\{tuple\_\\
delimiter\}<entity\_description>\\
\\
2. From the entities identified in step 1, identify all pairs of (source\_entity, target\_entity) that are *clearly related* to each other\\
For each pair of related entities, extract the following information:\\
- source\_entity: name of the source entity, as identified in step 1\\
- target\_entity: name of the target entity, as identified in step 1\\
- relationship\_description: explanation as to why you think the source entity and the target entity are related to each other\\
- relationship\_strength: a numeric score indicating strength of the relationship between the source entity and target entity\\
\\
Format each relationship as ("relationship"\{tuple\_delimiter\}<source\_entity>\{tuple\_delimiter\}<target\_\\
entity>\{tuple\_delimiter\}<relationship\_description>\{tuple\_delimiter\}<relationship\_strength>)\\
\\
3. Return output in English as a single list of all the entities and relationships identified in steps 1 and 2. Use **\{record\_delimiter\}** as the list delimiter.\\
\\
4. When finished, output \{completion\_delimiter\}\\
\\
---Examples---\\
\\
Entity\_types: ORGANIZATION,PERSON\\
\\
Input:\\
\\
The Fed is scheduled to meet on Tuesday and Wednesday, with the central bank planning to release its latest policy decision on Wednesday at 2:00 p.m. ET, followed by a press conference where Fed Chair Jerome Powell will take questions. Investors expect the Federal Open Market Committee to hold its benchmark interest rate steady in a range of 5.25\%-5.5\%.\\
\\
Output:\\
\\
("entity"\{tuple\_delimiter\}FED\{tuple\_delimiter\}ORGANIZATION\{tuple\_delimiter\}The Fed is the Federal Reserve, which is setting interest rates on Tuesday and Wednesday)\\
\{record\_delimiter\}\\
("entity"\{tuple\_delimiter\}JEROME POWELL\{tuple\_delimiter\}PERSON\{tuple\_delimiter\}Jerome Powell is the chair of the Federal Reserve)\\
\{record\_delimiter\}\\
("entity"\{tuple\_delimiter\}FEDERAL OPEN MARKET COMMITTEE\{tuple\_delimiter\}ORGANIZATION\{tuple\_delimiter\}The Federal Reserve committee makes key decisions about interest rates and the growth of the United States money supply)\\
\{record\_delimiter\}\\
("relationship"\{tuple\_delimiter\}JEROME POWELL\{tuple\_delimiter\}FED\{tuple\_delimiter\}Jerome Powell is the Chair of the Federal Reserve and will answer questions at a press conference\{tuple\_delimiter\}9)\\
\{completion\_delimiter\}\\
}
\textit{\\
...More examples...\\
\\}
\texttt{---Real Data---\\
\\
Entity\_types: \{entity\_types\}\\
Input:\\
\\
\{input\_text\}\\
\\
Output:}

\subsection{Community Summary Generation}

\texttt{---Role--- \\
You are an AI assistant that helps a human analyst to perform general information discovery. Information discovery is the process of identifying and assessing relevant information associated with certain entities (e.g., organizations and individuals) within a network.\\
\\
---Goal---\\
Write a comprehensive report of a community, given a list of entities that belong to the community as well as their relationships and optional associated claims. The report will be used to inform decision-makers about information associated with the community and their potential impact. The content of this report includes an overview of the community's key entities, their legal compliance, technical capabilities, reputation, and noteworthy claims.\\
\\
---Report Structure---\\
\\
The report should include the following sections:\\
\\
- TITLE: community's name that represents its key entities - title should be short but specific. When possible, include representative named entities in the title.\\
- SUMMARY: An executive summary of the community's overall structure, how its entities are related to each other, and significant information associated with its entities.\\
- IMPACT SEVERITY RATING: a float score between 0-10 that represents the severity of IMPACT posed by entities within the community.  IMPACT is the scored importance of a community.\\
- RATING EXPLANATION: Give a single sentence explanation of the IMPACT severity rating.\\
- DETAILED FINDINGS: A list of 5-10 key insights about the community. Each insight should have a short summary followed by multiple paragraphs of explanatory text grounded according to the grounding rules below. Be comprehensive.\\
\\
Return output as a well-formed JSON-formatted string with the following format:\\
    \{\{\\
        "title": <report\_title>,\\
        "summary": <executive\_summary>,\\
        "rating": <impact\_severity\_rating>,\\
        "rating\_explanation": <rating\_explanation>,\\
        "findings": [\\
            \{\{\\
                "summary":<insight\_1\_summary>,\\
                "explanation": <insight\_1\_explanation>\\
            \}\},\\
            \{\{\\
                "summary":<insight\_2\_summary>,\\
                "explanation": <insight\_2\_explanation>\\
            \}\}\\
        ]\\
    \}\}\\
\\
---Grounding Rules---\\
\\
Points supported by data should list their data references as follows:\\
\\
"This is an example sentence supported by multiple data references [Data: <dataset name> (record ids); <dataset name> (record ids)]."\\
\\
Do not list more than 5 record ids in a single reference. Instead, list the top 5 most relevant record ids and add "+more" to indicate that there are more.\\
\\
For example:\\
"Person X is the owner of Company Y and subject to many allegations of wrongdoing [Data: Reports (1), Entities (5, 7); Relationships (23); Claims (7, 2, 34, 64, 46, +more)]."\\
\\
where 1, 5, 7, 23, 2, 34, 46, and 64 represent the id (not the index) of the relevant data record.\\
\\
Do not include information where the supporting evidence for it is not provided.\\
\\
---Example---\\
\\
Input:\\
\\
Entities\\
\\
id,entity,description\\
5,VERDANT OASIS PLAZA,Verdant Oasis Plaza is the location of the Unity March\\
6,HARMONY ASSEMBLY,Harmony Assembly is an organization that is holding a march at Verdant Oasis Plaza\\
\\
Relationships\\
\\
id,source,target,description\\
37,VERDANT OASIS PLAZA,UNITY MARCH,Verdant Oasis Plaza is the location of the Unity March\\
38,VERDANT OASIS PLAZA,HARMONY ASSEMBLY,Harmony Assembly is holding a march at Verdant Oasis Plaza\\
39,VERDANT OASIS PLAZA,UNITY MARCH,The Unity March is taking place at Verdant Oasis Plaza\\
40,VERDANT OASIS PLAZA,TRIBUNE SPOTLIGHT,Tribune Spotlight is reporting on the Unity march taking place at Verdant Oasis Plaza\\
41,VERDANT OASIS PLAZA,BAILEY ASADI,Bailey Asadi is speaking at Verdant Oasis Plaza about the march\\
43,HARMONY ASSEMBLY,UNITY MARCH,Harmony Assembly is organizing the Unity March\\
\\
Output:\\
\{\{\\
    "title": "Verdant Oasis Plaza and Unity March",\\
    "summary": "The community revolves around the Verdant Oasis Plaza, which is the location of the Unity March. The plaza has relationships with the Harmony Assembly, Unity March, and Tribune Spotlight, all of which are associated with the march event.",\\
    "rating": 5.0,\\
    "rating\_explanation": "The impact severity rating is moderate due to the potential for unrest or conflict during the Unity March.",\\
    "findings": [\\
        \{\{\\
            "summary": "Verdant Oasis Plaza as the central location",\\
            "explanation": "Verdant Oasis Plaza is the central entity in this community, serving as the location for the Unity March. This plaza is the common link between all other entities, suggesting its significance in the community. The plaza's association with the march could potentially lead to issues such as public disorder or conflict, depending on the nature of the march and the reactions it provokes. [Data: Entities (5), Relationships (37, 38, 39, 40, 41,+more)]"\\
        \}\},\\
        \{\{\\
            "summary": "Harmony Assembly's role in the community",\\
            "explanation": "Harmony Assembly is another key entity in this community, being the organizer of the march at Verdant Oasis Plaza. The nature of Harmony Assembly and its march could be a potential source of threat, depending on their objectives and the reactions they provoke. The relationship between Harmony Assembly and the plaza is crucial in understanding the dynamics of this community. [Data: Entities(6), Relationships (38, 43)]"\\
        \}\},\\
        \{\{\\
            "summary": "Unity March as a significant event",\\
            "explanation": "The Unity March is a significant event taking place at Verdant Oasis Plaza. This event is a key factor in the community's dynamics and could be a potential source of threat, depending on the nature of the march and the reactions it provokes. The relationship between the march and the plaza is crucial in understanding the dynamics of this community. [Data: Relationships (39)]"\\
        \}\},\\
        \{\{\\
            "summary": "Role of Tribune Spotlight",
            "explanation": "Tribune Spotlight is reporting on the Unity March taking place in Verdant Oasis Plaza. This suggests that the event has attracted media attention, which could amplify its impact on the community. The role of Tribune Spotlight could be significant in shaping public perception of the event and the entities involved. [Data: Relationships (40)]"\\
        \}\}\\
    ]\\
\}\}\\
\\
---Real Data---\\
\\
Use the following text for your answer. Do not make anything up in your answer.\\
\\
Input:\\
\\
\{input\_text\}\\}
\textit{\\
...Report Structure and Grounding Rules Repeated...\\
\\}
\texttt{Output:}

\subsection{Community Answer Generation}

\texttt{---Role---\\
\\
You are a helpful assistant responding to questions about a dataset by synthesizing perspectives from multiple analysts.\\
\\
---Goal---\\
\\
Generate a response of the target length and format that responds to the user's question, summarize all the reports from multiple analysts who focused on different parts of the dataset, and incorporate any relevant general knowledge.\\
\\
Note that the analysts' reports provided below are ranked in the **descending order of helpfulness**.\\
\\
If you don't know the answer, just say so. Do not make anything up.\\
\\
The final response should remove all irrelevant information from the analysts' reports and merge the cleaned information into a comprehensive answer that provides explanations of all the key points and implications appropriate for the response length and format.\\
\\
Add sections and commentary to the response as appropriate for the length and format. Style the response in markdown.\\
\\
The response shall preserve the original meaning and use of modal verbs such as "shall", "may" or "will".\\
\\
The response should also preserve all the data references previously included in the analysts' reports, but do not mention the roles of multiple analysts in the analysis process.\\
\\
Do not list more than 5 record ids in a single reference. Instead, list the top 5 most relevant record ids and add "+more" to indicate that there are more.\\
\\
For example:\\
\\
"Person X is the owner of Company Y and subject to many allegations of wrongdoing [Data: Reports (2, 7, 34, 46, 64, +more)]. He is also CEO of company X [Data: Reports (1, 3)]"\\
\\
where 1, 2, 3, 7, 34, 46, and 64 represent the id (not the index) of the relevant data record.\\
\\
Do not include information where the supporting evidence for it is not provided.\\
\\
---Target response length and format---\\
\\
\{response\_type\}\\
\\
---Analyst Reports---\\
\\
\{report\_data\}\\
\\}
\textit{\\
...Goal and Target response length and format repeated...\\
\\}
\texttt{Add sections and commentary to the response as appropriate for the length and format. Style the response in markdown.\\
\\
Output:}

\subsection{Global Answer Generation}

\texttt{---Role---\\
\\
You are a helpful assistant responding to questions about data in the tables provided.\\
\\
---Goal---\\
\\
Generate a response of the target length and format that responds to the user's question, summarize all relevant information in the input data tables appropriate for the response length and format, and incorporate any relevant general knowledge.\\
\\
If you don't know the answer, just say so. Do not make anything up.\\
\\
The response shall preserve the original meaning and use of modal verbs such as "shall", "may" or "will".\\
\\
Points supported by data should list the relevant reports as references as follows:\\
\\
"This is an example sentence supported by data references [Data: Reports (report ids)]"\\
\\}
\textit{Note: the prompts for SS (semantic search) and TS (text summarization) conditions use "Sources" in place of "Reports" above.\\}
\texttt{\\
Do not list more than 5 record ids in a single reference. Instead, list the top 5 most relevant record ids and add "+more" to indicate that there are more.\\
\\
For example:\\
\\
"Person X is the owner of Company Y and subject to many allegations of wrongdoing [Data: Reports (2, 7, 64, 46, 34, +more)]. He is also CEO of company X [Data: Reports (1, 3)]"\\
\\
where 1, 2, 3, 7, 34, 46, and 64 represent the id (not the index) of the relevant data report in the provided tables.\\
\\
Do not include information where the supporting evidence for it is not provided.\\
\\
At the beginning of your response, generate an integer score between 0-100 that indicates how **helpful** is this response in answering the user's question. Return the score in this format: <ANSWER\_HELPFULNESS> score\_value </ANSWER\_HELPFULNESS>.\\
\\
---Target response length and format---\\
\\
\{response\_type\}\\
\\
---Data tables---\\
\\
\{context\_data\}\\}
\textit{\\
...Goal and Target response length and format repeated...\\
\\}
Output:

\raggedbottom

\pagebreak
\section{Evaluation Prompts}
\label{app:eval_prompts}
\subsection{Relative Assessment Prompt}

\texttt{---Role---\\
\\
You are a helpful assistant responsible for grading two answers to a question that are provided by two different people.\\
\\
---Goal---\\
\\
Given a question and two answers (Answer 1 and Answer 2), assess which answer is better according to the following measure:\\
\\
\{criteria\}\\
\\
Your assessment should include two parts:\\
- Winner: either 1 (if Answer 1 is better) and 2 (if Answer 2 is better) or 0 if they are fundamentally similar and the differences are immaterial.\\
- Reasoning: a short explanation of why you chose the winner with respect to the measure described above.\\
\\
Format your response as a JSON object with the following structure:\\
\{\{\\
    "winner": <1, 2, or 0>,\\
    "reasoning": "Answer 1 is better because <your reasoning>."\\
\}\}\\
\\
---Question---\\
\\
\{question\}\\
\\
---Answer 1---\\
\\
\{answer1\}\\
\\
---Answer 2---\\
\\
\{answer2\}\\
\\
Assess which answer is better according to the following measure:\\
\\
\{criteria\}\\
\\
Output:}

\subsection{Relative Assessment Metrics}

\texttt{CRITERIA = \{\\
    "comprehensiveness": "How much detail does the answer provide to cover all the aspects and details of the question? A comprehensive answer should be thorough and complete, without being redundant or irrelevant. For example, if the question is 'What are the benefits and drawbacks of nuclear energy?', a comprehensive answer would provide both the positive and negative aspects of nuclear energy, such as its efficiency, environmental impact, safety, cost, etc. A comprehensive answer should not leave out any important points or provide irrelevant information. For example, an incomplete answer would only provide the benefits of nuclear energy without describing the drawbacks, or a redundant answer would repeat the same information multiple times.",\\
    "diversity": "How varied and rich is the answer in providing different perspectives and insights on the question? A diverse answer should be multi-faceted and multi-dimensional, offering different viewpoints and angles on the question. For example, if the question is 'What are the causes and effects of climate change?', a diverse answer would provide different causes and effects of climate change, such as greenhouse gas emissions, deforestation, natural disasters, biodiversity loss, etc. A diverse answer should also provide different sources and evidence to support the answer. For example, a single-source answer would only cite one source or evidence, or a biased answer would only provide one perspective or opinion.",\\
    "directness": "How specifically and clearly does the answer address the question? A direct answer should provide a clear and concise answer to the question. For example, if the question is 'What is the capital of France?', a direct answer would be 'Paris'. A direct answer should not provide any irrelevant or unnecessary information that does not answer the question. For example, an indirect answer would be 'The capital of France is located on the river Seine'.",\\
    "empowerment": "How well does the answer help the reader understand and make informed judgements about the topic without being misled or making fallacious assumptions. Evaluate each answer on the quality of answer as it relates to clearly explaining and providing reasoning and sources behind the claims in the answer."\\
\}}

\raggedbottom

\pagebreak
\section{Statistical Analysis}
\label{app:stats} 

\begin{table}[htp]
\tiny
\centering
\caption{Pairwise comparisons of six conditions on four metrics across 125 questions and two datasets. For each question and metric, the winning condition received a score of 100, the losing condition received a score of 0, and in the event of a tie, each condition was scored 50. These scores were then averaged over five evaluation runs for each condition. Results of Shapiro-Wilk tests indicated that the data did not follow a normal distribution. Thus, non-parametric tests (Wilcoxon signed-rank tests) were employed to assess the performance differences between pairs of conditions, with Holm-Bonferroni correction applied to account for multiple pairwise comparisons. The corrected p-values that indicated statistically significant differences are highlighted in bold.}
\begin{tabular}{l l l l l l l l l l l}  
\hline
& & & \multicolumn{4}{l}{\textbf{Podcast Transcripts}} & \multicolumn{4}{l}{\textbf{News Articles}} \\
\hline
& \textbf{Condition 1} & \textbf{Condition 2} & \textbf{Mean 1} & \textbf{Mean 2} & \textbf{Z-value} & \textbf{p-value} & \textbf{Mean 1} & \textbf{Mean 2} & \textbf{Z-value} & \textbf{p-value} \\
 \hline
\multirow{15}{*}{\textbf{Comprehensiveness}} & C0 & TS & 50.24 & 49.76 & -0.06 & 1
& 55.52 & 44.48 & -2.03 & 0.17
\\
 & C1 & TS & 51.92 & 48.08 & -1.56 & 0.633
& 58.8 & 41.2 & -3.62 &\textbf{ 0.002}
\\
 & C2 & TS & 57.28 & 42.72 & -4.1 & \textbf{\textless0.001}
& 62.08 & 37.92 & -5.07 &\textbf{\textless0.001}
\\
 & C3 & TS & 56.48 & 43.52 & -3.42 &\textbf{ 0.006}
& 63.6 & 36.4 & -5.63 & \textbf{\textless0.001}
\\
 & C0 & SS & 71.92 & 28.08 & -6.2 & \textbf{\textless0.001}
& 71.76 & 28.24 & -6.3 & \textbf{\textless0.001}
\\
 & C1 & SS & 75.44 & 24.56 & -7.45 &\textbf{\textless0.001}
& 74.72 & 25.28 & -7.78 & \textbf{\textless0.001}
\\
 & C2 & SS & 77.76 & 22.24 & -8.17 & \textbf{\textless0.001}
& 79.2 & 20.8 & -8.34 & \textbf{\textless0.001}
\\
 & C3 & SS & 78.96 & 21.04 & -8.12 & \textbf{\textless0.001}
& 79.44 & 20.56 & -8.44 & \textbf{\textless0.001}
\\
 & TS & SS & 83.12 & 16.88 & -8.85 & \textbf{\textless0.001}
& 79.6 & 20.4 & -8.27 & \textbf{\textless0.001}
\\
 & C0 & C1 & 53.2 & 46.8 & -1.96 & 0.389
& 51.92 & 48.08 & -0.45 & 0.777
\\
 & C0 & C2 & 50.24 & 49.76 & -0.23 & 1
& 53.68 & 46.32 & -1.54 & 0.371
\\
 & C1 & C2 & 51.52 & 48.48 & -1.62 & 0.633
& 57.76 & 42.24 & -4.01 & \textbf{\textless0.001}
\\
 & C0 & C3 & 49.12 & 50.88 & -0.56 & 1
& 52.16 & 47.84 & -0.86 & 0.777
\\
 & C1 & C3 & 50.32 & 49.68 & -0.66 & 1
& 55.12 & 44.88 & -2.94 &\textbf{0.016}
\\
 & C2 & C3 & 52.24 & 47.76 & -1.97 & 0.389
& 58.64 & 41.36 & -3.68 & \textbf{0.002}
\\
 \hline
\multirow{15}{*}{\textbf{Diversity}} & C0 & TS & 50.24 & 49.76 & -0.11 & 1
& 46.88 & 53.12 & -1.38 & 0.676
\\
 & C1 & TS & 50.48 & 49.52 & -0.12 & 1
& 54.64 & 45.36 & -1.88 & 0.298
\\
 & C2 & TS & 57.12 & 42.88 & -2.84 & \textbf{0.036}
& 55.76 & 44.24 & -2.16 & 0.184
\\
 & C3 & TS & 54.32 & 45.68 & -2.39 & 0.1
& 60.16 & 39.84 & -4.07 & \textbf{\textless0.001}
\\
 & C0 & SS & 76.56 & 23.44 & -7.12 & \textbf{\textless0.001}
& 62.08 & 37.92 & -3.57 & \textbf{0.003}
\\
 & C1 & SS & 75.44 & 24.56 & -7.33 & \textbf{\textless0.001}
& 64.96 & 35.04 & -4.92 & \textbf{\textless0.001}
\\
 & C2 & SS & 80.56 & 19.44 & -8.21 & \textbf{\textless0.001}
& 70.56 & 29.44 & -6.29 & \textbf{\textless0.001}
\\
 & C3 & SS & 80.8 & 19.2 & -8.3 & \textbf{\textless0.001}
& 69.12 & 30.88 & -5.53 & \textbf{\textless0.001}
\\
 & TS & SS & 82.08 & 17.92 & -8.43 & \textbf{\textless0.001}
& 67.2 & 32.8 & -4.85 & \textbf{\textless0.001}
\\
 & C0 & C1 & 49.76 & 50.24 & -0.13 & 1
& 39.68 & 60.32 & -3.61 &\textbf{0.003}
\\
 & C0 & C2 & 46.32 & 53.68 & -1.5 & 0.669
& 40.96 & 59.04 & -3.14 & \textbf{0.012}
\\
 & C1 & C2 & 44.08 & 55.92 & -3.27 & \textbf{0.011}
& 50.24 & 49.76 & -0.22 & 1
\\
 & C0 & C3 & 44 & 56 & -2.6 & 0.065
& 41.04 & 58.96 & -3.47 & \textbf{0.004}
\\
 & C1 & C3 & 45.44 & 54.56 & -2.98 & \textbf{0.026}
& 49.52 & 50.48 & -0.01 & 1
\\
 & C2 & C3 & 48.48 & 51.52 & -0.96 & 1
& 50.96 & 49.04 & -0.39 & 1
\\
 \hline
\multirow{15}{*}{\textbf{Empowerment}} & C0 & TS & 40.96 & 59.04 & -4.3 & \textbf{\textless0.001}
& 42.24 & 57.76 & -3.32 & \textbf{0.012}
\\
 & C1 & TS & 45.2 & 54.8 & -3.76 & \textbf{0.002}
& 50 & 50 & -0.12 & 1
\\
 & C2 & TS & 47.68 & 52.32 & -2.2 & 0.281
& 49.52 & 50.48 & -0.22 & 1
\\
 & C3 & TS & 48.72 & 51.28 & -1.27 & 1
& 51.68 & 48.32 & -1.2 & 1
\\
 & C0 & SS & 42.96 & 57.04 & -3.71 & \textbf{0.003}
& 42.72 & 57.28 & -3.12 & \textbf{0.022}
\\
 & C1 & SS & 47.68 & 52.32 & -1.5 & 0.936
& 51.36 & 48.64 & -0.84 & 1
\\
 & C2 & SS & 50.72 & 49.28 & -0.55 & 1
& 49.84 & 50.16 & -0.2 & 1
\\
 & C3 & SS & 48.96 & 51.04 & -0.57 & 1
& 49.52 & 50.48 & -0.08 & 1
\\
 & TS & SS & 57.52 & 42.48 & -4.1 &\textbf{ \textless0.001}
& 52.88 & 47.12 & -1.1 & 1
\\
 & C0 & C1 & 48.72 & 51.28 & -1.23 & 1
& 42.4 & 57.6 & -3.9 & \textbf{0.001}
\\
 & C0 & C2 & 46.64 & 53.36 & -2.54 & 0.12
& 44.8 & 55.2 & -2.16 & 0.336
\\
 & C1 & C2 & 49.28 & 50.72 & -1.73 & 0.682
& 52 & 48 & -1.45 & 1
\\
 & C0 & C3 & 47.6 & 52.4 & -1.78 & 0.682
& 44.32 & 55.68 & -3.45 & \textbf{0.008}
\\
 & C1 & C3 & 50 & 50 & 0 & 1
& 51.44 & 48.56 & -1.02 & 1
\\
 & C2 & C3 & 50.72 & 49.28 & -0.86 & 1
& 50.4 & 49.6 & -0.22 & 1
\\
 \hline
\multirow{15}{*}{\textbf{Directness}} & C0 & TS & 44.96 & 55.04 & -4.09 & \textbf{\textless0.001}
& 45.2 & 54.8 & -3.68 & \textbf{0.003}
\\
 & C1 & TS & 47.92 & 52.08 & -2.41 & 0.126
& 46.64 & 53.36 & -2.91 & \textbf{0.04}
\\
 & C2 & TS & 48.8 & 51.2 & -2.23 & 0.179
& 48.32 & 51.68 & -2.12 & 0.179
\\
 & C3 & TS & 48.08 & 51.92 & -2.23 & 0.179
& 48.32 & 51.68 & -2.56 & 0.074
\\
 & C0 & SS & 35.12 & 64.88 & -6.17 &\textbf{ \textless0.001}
& 41.44 & 58.56 & -4.82 & \textbf{\textless0.001}
\\
 & C1 & SS & 40.32 & 59.68 & -4.83 & \textbf{\textless0.001}
& 45.2 & 54.8 & -3.19 & \textbf{0.017}
\\
 & C2 & SS & 40.4 & 59.6 & -4.67 & \textbf{\textless0.001}
& 44.88 & 55.12 & -3.65 & \textbf{0.003}
\\
 & C3 & SS & 40.48 & 59.52 & -4.69 & \textless\textbf{0.001}
& 45.6 & 54.4 & -2.86 & \textbf{0.043}
\\
 & TS & SS & 43.6 & 56.4 & -3.96 &\textbf{\textless0.001}
& 46 & 54 & -2.68 & 0.066
\\
 & C0 & C1 & 46.96 & 53.04 & -2.87 & \textbf{0.037}
& 47.6 & 52.4 & -2.17 & 0.179
\\
 & C0 & C2 & 48.4 & 51.6 & -2.06 & 0.197
& 48.48 & 51.52 & -1.61 & 0.321
\\
 & C1 & C2 & 49.84 & 50.16 & -1 & 0.952
& 49.28 & 50.72 & -1.6 & 0.321
\\
 & C0 & C3 & 48.4 & 51.6 & -1.8 & 0.29
& 47.2 & 52.8 & -2.62 & 0.071
\\
 & C1 & C3 & 49.76 & 50.24 & 0 & 1
& 48.8 & 51.2 & -1.29 & 0.321
\\
 & C2 & C3 & 50 & 50 & 0 & 1
& 48.8 & 51.2 & -1.84 & 0.262
\\

\end{tabular}
\label{tab:stats_analysis}
\end{table}

\raggedbottom

\end{document}